\documentclass[10pt,twocolumn,letterpaper]{article}

\usepackage{cvpr}
\usepackage{times}
\usepackage{epsfig}
\usepackage{graphicx}
\usepackage{amsmath}
\usepackage{amssymb}
\usepackage{multirow}
\usepackage{booktabs}
\usepackage{array}
\usepackage{subfigure}
\usepackage{tabularx}
\usepackage{bbm}
\usepackage{pifont}
\usepackage{xcolor}
\usepackage{titling}

\usepackage[pagebackref=true,breaklinks=true,letterpaper=true,colorlinks,bookmarks=false]{hyperref}

\cvprfinalcopy 

\def\cvprPaperID{} 

\ifcvprfinal\fi
\begin{document}

\title{\textbf{DecAug: Augmenting HOI Detection via Decomposition}}

\author{Hao-Shu Fang$^{1*}$, Yichen Xie$^{1*}$, Dian Shao$^2$, Yong-Lu Li$^{1}$, Cewu Lu$^{1\dagger}$\\
$^1$Shanghai Jiao Tong University \quad
$^2$The Chinese University of Hong Kong\\
{\tt\small fhaoshu@gmail.com, xieyichen@sjtu.edu.cn,} \\
{\tt\small sd017@ie.cuhk.edu.hk, yonglu\_li@sjtu.edu.cn, lucewu@sjtu.edu.cn}
}

\date{}
\maketitle
\let\thefootnote\relax\footnotetext{$*$ Equal contribution. Names in alphabetical order.}
\let\thefootnote\relax\footnotetext{$\dagger$ Cewu Lu is the corresponding author.}

\begin{abstract}
Human-object interaction (HOI) detection requires a large amount of annotated data. Current algorithms suffer from insufficient training samples and category imbalance within datasets. To increase data efficiency, in this paper, we propose an efficient and effective data augmentation method called \textbf{DecAug} for HOI detection. Based on our proposed object state similarity metric, object patterns across different HOIs are shared to augment local object appearance features without changing their state. Further, we shift spatial correlation between humans and objects to other feasible configurations with the aid of a pose-guided Gaussian Mixture Model while preserving their interactions. Experiments show that our method brings up to \textbf{3.3 mAP} and \textbf{1.6 mAP} improvements on V-COCO and HICO-DET dataset for two advanced models. Specifically, interactions with fewer samples enjoy more notable improvement. Our method can be easily integrated into various HOI detection models with negligible extra computational consumption. Our code will be made publicly available.
\end{abstract}

\begin{figure}[t]
\centering
\subfigure[Instance-Level Augmentation Example: \textit{heatmap-guided instaboost}~\cite{fang2019instaboost}  (left: original, right: augmented)]{
\includegraphics[width=0.45\linewidth]{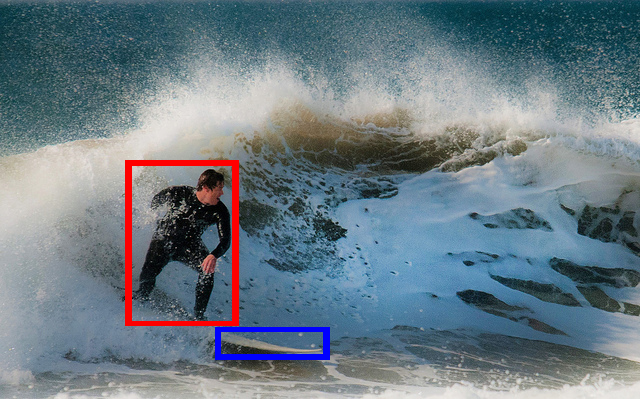}
\includegraphics[width=0.45\linewidth]{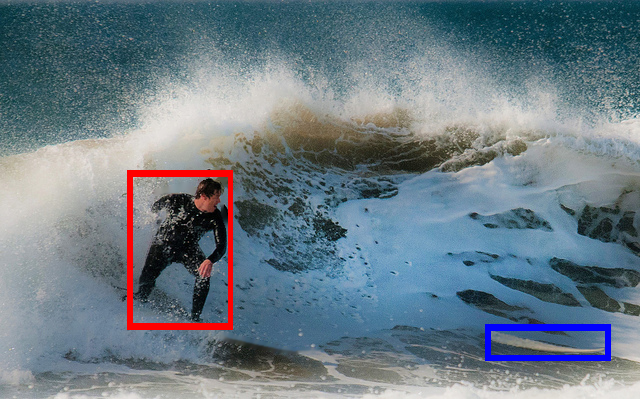}
\label{fig:drawback(a)}
}
\subfigure[Our Approach: local object appearance augmentation (left) and global spatial correlation augmentation (right)]{
\includegraphics[width=0.45\linewidth]{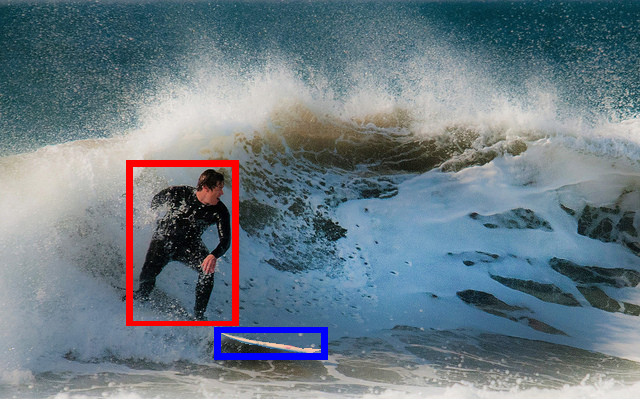}
\includegraphics[width=0.45\linewidth]{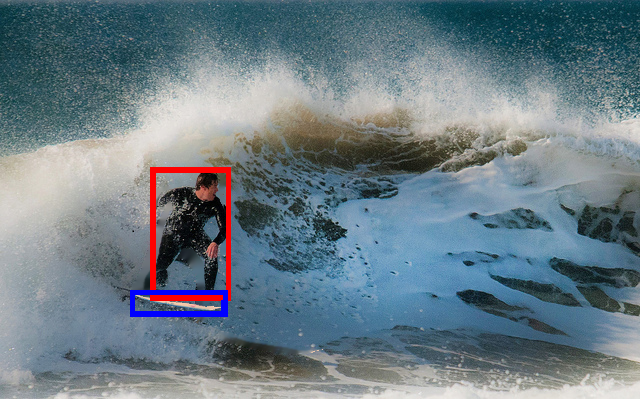}
\label{fig:drawback(b)}
}

\caption{(a) shows the result of \textit{heatmap-guided instaboost}. The left is the original image while the right has been augmented. The board is moved far away, which has no negative effect on object detection or instance segmentation. However, it devastates the relationship between the human and object. In contrast, (b) shows the two steps of our DecAug. Local object appearance is changed in the left image. Then, global spatial correlation augmentation is applied in the right one. The human-object interaction remains distinguishable in both.
}
\label{fig:drawback}
\end{figure}

\section{Introduction}
Human-object interaction (HOI) detection aims to localize humans and objects as well as infer their interaction categories in a still image. For each interaction, a triplet of $\langle subject, predicate, object\rangle$ should be retrieved. As a sub-task of visual relationship detection, HOI detection pays attention to human-centric interactions with objects. It plays an essential role in the understanding of scenes, which facilitates many other fields like activity understanding~\cite{caba2015activitynet},  image captioning~\cite{li2017scene} and robot learning~\cite{argall2009survey}.

Along with the recent achievements computer vision has reached, many exciting deep neural network (DNN) models for HOI detection have been developed. They took various types of features into account such as visual features~\cite{gupta2015visual}, spatial location~\cite{chao2018learning,xu2019interact}, human poses~\cite{yao2010modeling,Gkioxari_2018_CVPR} or text corpus~\cite{liang2020visual}. However, the progress of HOI detection is still slower compared with the achievement in other tasks like object detection and instance segmentation. There are currently two main hindrances to further performance gains. For one thing, HOI detection depends on a better understanding of contextual information. It calls for a large amount of high quality data. However, large datasets are not easily accessible due to the labor intensity of annotation. For another thing, an apparent imbalance inevitably exists between different interaction categories in current large datasets~\cite{gupta2015visual,chao2018learning,zhuang2017care}. Some interactions naturally have much more positive samples than others, such as \textit{look at}, \textit{sit on} and \textit{stand on}, which causes a serious long-tail issue.

To tackle such problems, a natural idea is to resort to data augmentation, whose power has been witnessed in many other tasks of computer vision~\cite{cubuk2019randaugment,simard2003best,hinterstoisser2019annotation,liu2016ssd,peng2018jointly,jaderberg2015spatial,fang2019instaboost}.
Unfortunately, previous research in cognition~\cite{baldassano2017human} demonstrated the difficulty of data augmentation for the task of HOI detection. Specifically, image-level random cropping cannot improve the diversity of interactions while instance movement damages the spatial correlation between humans and objects. As shown in Fig.~\ref{fig:drawback}(a), it is hard to identify the interaction in the images using such simple augmentation.


In this paper, we propose a novel data augmentation method named \textbf{DecAug}. 
Aiming to improve the diversity of interactions without semantic loss, \textbf{DecAug} mainly includes two components: \textit{local object appearance augmentation} and \textit{global spatial correlation augmentation}.

To elaborate, for local object appearance, we propose a simple but effective cross-image instance substitution technique to increase the generalization ability of 
models towards entity concepts instead of object patterns. An object state similarity metric is also introduced to 
justify the replacement of an object with another based on their state coherency.

Furthermore, we try to augment the global spatial correlations between humans and objects without contextual loss. According to \cite{knill1996introduction}, the perceptual inference of human derives from information available to observers and some \emph{empirical knowledge} of the world. Intuitively, reasonable placement of objects could also be obtained with prior knowledge from the whole dataset. Inspired by the strong correlation between human pose and HOI~\cite{yao2012recognizing}, we build a probability distribution of object location for each training sample, which comes from the spatial relationship of other samples with similar human poses. With this distribution aware augmentation, we are able to improve the diversity within each interaction without damaging their semantic meanings.

We conduct extensive experiments on two mainstream datasets: V-COCO~\cite{gupta2015visual} and HICO-DET~\cite{chao2018learning}. After augmentation, the performance of two advanced open-source models (iCAN~\cite{gao2018ican} and Transferable Interactiveness Network~\cite{li2019transferable}) can be improved by a large margin (\textbf{3.3} and \textbf{2.6 mAP} on V-COCO; \textbf{1.6} and \textbf{1.3 mAP} on HICO-DET). Same object detection proposals are used to ensure the improvements come from interaction recognition instead of object detection. Specifically, for those interactions with fewer positive samples, the improvement is more notable, suggesting our method helps tackle the long-tail issue. Our code will be made publicly available.

\section{Related Work}
\subsection{Visual Relationship Detection}
Visual relationship detection~\cite{lu2016visual,xu2017scene,gkioxari2018detecting,zellers2018neural,zhang2017visual} needs to not only find objects location in an image but also detect the relationships between them. These relationships includes actions~\cite{Shao_2020_CVPR}, interactions~\cite{gkioxari2018detecting} or other more general relationships~\cite{lu2016visual,zhang2017visual}. Different from object detection or instance segmentation, visual relationship detection requires to exploit more semantic information~\cite{baldassano2017human} like the spatial positions of humans and objects~\cite{chao2018learning}. Since such semantic information is difficult to extract, enough training samples are necessary for these models. Requirement for maintaining the semantic information also poses an extra challenge to data augmentation.

\subsection{Human-Object Interaction Detection}
Human-object interaction (HOI) detection task is significant for understanding human behavior with objects. Some early work~\cite{gupta2015visual} tried to detect humans and objects separately, which led to limited performance. Christopher \textit{et. al.}~\cite{baldassano2017human} proposed that rather than the sum of parts, more information should be taken into consideration. Gao \textit{et. al.}~\cite{gao2018ican} proposed an instance-centric attention module to enhance regions of interest. Chao \textit{et. al.}~\cite{chao2018learning} added the relative spatial relationship between humans and objects into the input of CNN. The significance of pair spatial configuration was further emphasized by Ulutan \textit{et.al.} and Wang \textit{et.al.} \cite{ulutan2020vsgnet, Wang_2020_CVPR}, which helped associate the interacted humans and objects. Some recent works~\cite{fang2018pairwise, wan2019pose,qi2018learning, li2019transferable, li2020pastanet} also thought of human poses as a crucial indicator of interaction.

More information means a higher requirement for data amount. There exist some popular datasets for this task such as V-COCO~\cite{gupta2015visual}, HICO-DET~\cite{chao2018learning}, HAKE~\cite{li2019hake} and HCVRD~\cite{zhuang2017care}. However, these datasets suffer from internal imbalance between different interaction categories, which is the so-called long-tail issue. Some interaction categories lack positive samples, which encumbers the overall performance. By increasing the diversity of data, data augmentation may help to solve this problem.

\subsection{Data Augmentation}
Data augmentation has been widely used in many tasks in the field of computer vision, such as image classification~\cite{cubuk2019randaugment,simard2003best,krizhevsky2012imagenet}, object detection~\cite{hinterstoisser2019annotation,liu2016ssd}, and pose estimation~\cite{peng2018jointly}. By generating additional training data, these methods helped to improve performance of various data-hungry models. Specifically, one branch of data augmentation focused on the instance-level, which fully exploited the fine-annotated segmentation of instances. Transformation applied on instances included scaling, rotation~\cite{jaderberg2015spatial}, jitter~\cite{fang2019instaboost}, pasting~\cite{kisantal2019augmentation} and affine transform~\cite{khoreva2019lucid}. However, all these above just utilized the information in a single image instead of the whole dataset. Some other work~\cite{choi2019self,qi2018semi,liu2017unsupervised} generated new images with Generative Adversarial Networks (GAN). Despite the impressive results, GAN needs plentiful extra training data, which is not applicable for current HOI datasets.

Another challenge rises about the placement of segmented instances on augmented images. Dvornik \textit{et. al.}~\cite{dvornik2018modeling} placed objects on the background according to the context. However, extra model needed to be trained beforehand. Fang \textit{et. al.}~\cite{fang2019instaboost} replaced the offline trained model with online context comparison. Yet, such a method does not preserve the visual relation information between instances inside an image.

Due to the difficulty in context preservation, there exists no effective data augmentation approach to generate extra training samples for visual relation detection tasks. Some prior effort~\cite{bansal2019detecting,hou2020visual} generated new interaction patterns based on word embedding but these could hardly improve visual diversity in training samples. In contrast, we develop a novel data augmentation method to visually boost data diversity for HOI detection. It makes use of information across the whole dataset as well as reserves visual relationships between humans and objects.

\begin{figure*}[htb!]
\centering

\subfigure[skateboard with the board]{
\includegraphics[width=3.6cm,height=2.4cm]{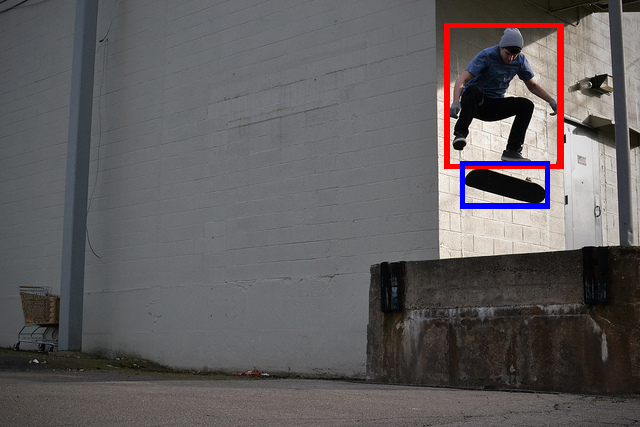}
\includegraphics[width=3.6cm,height=2.4cm]{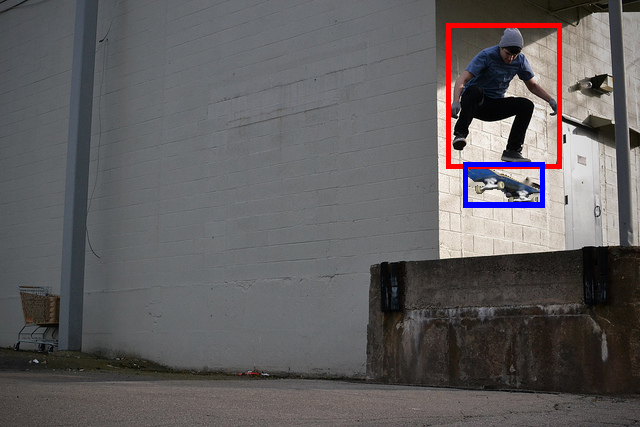}
\includegraphics[width=3.6cm,height=2.4cm]{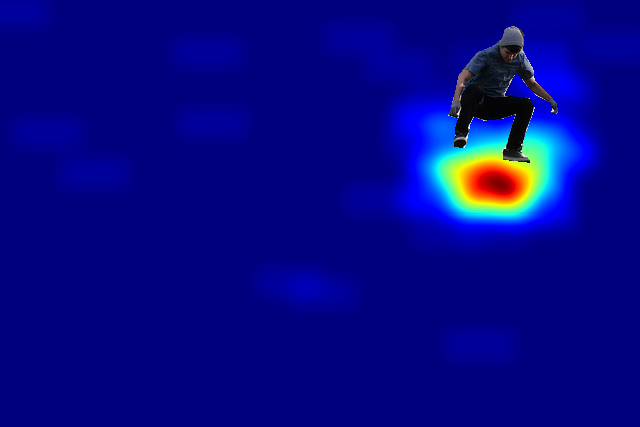}
\includegraphics[width=3.6cm,height=2.4cm]{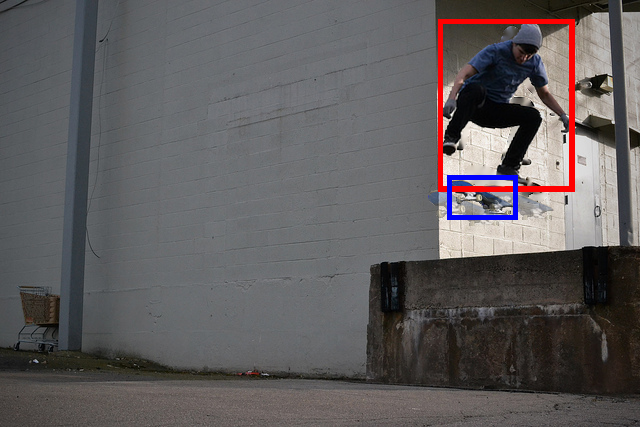}
\label{fig:overview(a)}
}

\subfigure[kick/look the ball]{
\includegraphics[width=3.6cm,height=2.4cm]{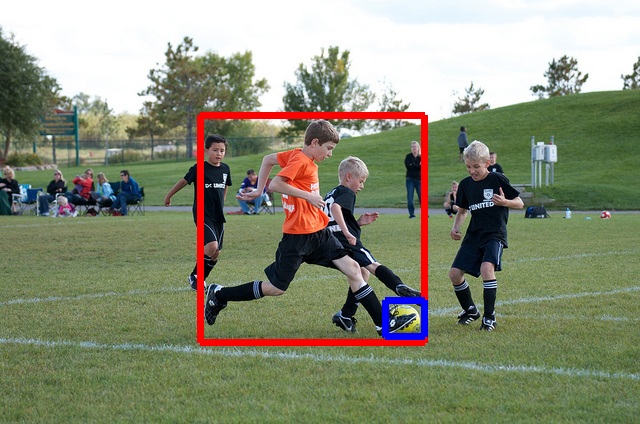}
\includegraphics[width=3.6cm,height=2.4cm]{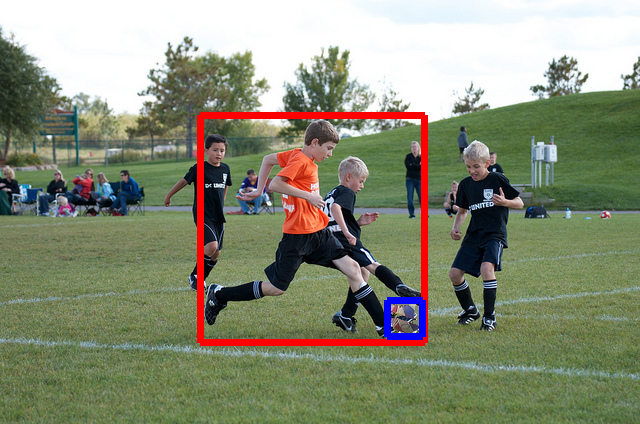}
\includegraphics[width=3.6cm,height=2.4cm]{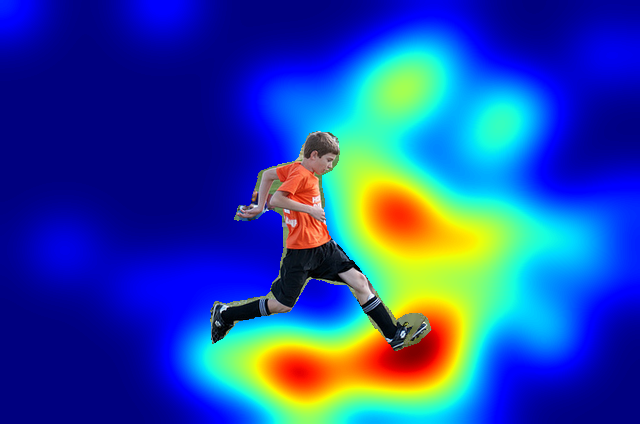}
\includegraphics[width=3.6cm,height=2.4cm]{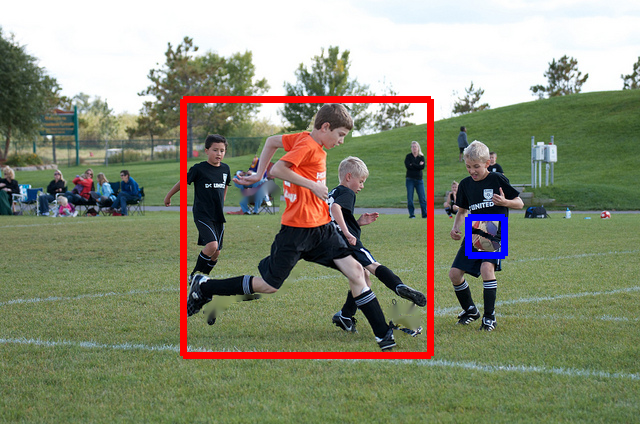}
}
\caption{Overview of our method: the first image is the original input (red box: human, blue box: object). The second image is the result of local object appearance augmentation (Sec.~\ref{sec:object exchange}). The third and forth images show the \textit{pose-guided probability distribution map} and the result of global spatial correlation augmentation (Sec.~\ref{sec:pose-guided}). Multiple translucent pastes are visualized in the rightest images to reflect the randomness.}
\label{fig:overview}
\end{figure*}

\section{Methods}
\subsection{Overview}
For the task of human-object interaction detection, we need to identify the interacting human-object pair, localize their positions and recognize their interaction category. In this paper, we focus on the interaction identification and recognition parts. Given detected humans and objects, a classifier $f$ needs to capture the very subtle details in the image to recognize the relationship $R$. A human-object interaction can be decomposed into the \textbf{background I}, the \textbf{human state h} including human appearance, pose, parsing, shape, gaze, etc., the \textbf{object state o} including category, 6D pose, occlusion, functionality, etc., and the \textbf{spatial relationship s} between the human and object. Mathematically, we have
\begin{equation}
    R = f(\textbf{I}, \textbf{h}, \textbf{o}, \textbf{s}).
\end{equation}
In this paper, we mainly augment the object state and spatial correlations, 
coherent with the human perception process.
This is nontrivial, since $R$ is very sensitive to the object state and spatial relations. We must find a manifold space in pixel level that could augment the object appearance while preserving the object state. In Sec.~\ref{sec:object exchange}, we introduce our local object appearance augmentation 
where an \textit{object state similarity metric} is proposed. Meanwhile, to find feasible spatial configurations for global spatial correlation augmentation, we propose the \textit{pose-guided probability distribution map} in Sec.~\ref{sec:pose-guided}. An overview of our method is shown in Fig.~\ref{fig:overview}.

\subsection{Local Object Appearance Augmentation}
\label{sec:object exchange}
When recognizing the HOI, the state of an object is far more important than its texture pattern. For example, when identifying the interaction of \textit{holding a mug}, the standing pose and the occlusion with hands are more important than the mug's color and texture. Thus, we propose to augment the local object appearance features to improve the generalization ability of the network, helping it pay more attention to the crucial object state instead of appearance. The key of such augmentation is to preserve the object state as much as possible. Meanwhile, patterns of augmented objects should be photo-realistic to avoid too many artifacts. Naturally, we can utilize the same category objects from the dataset during training \text{i.e.} we replace the object with suitable same category instances in other images. We then explain our principle for objects appearance replacement as follows.

\begin{figure}[tb!]
\centering
\subfigure{
\includegraphics[width=2.4cm,height=1.6cm]{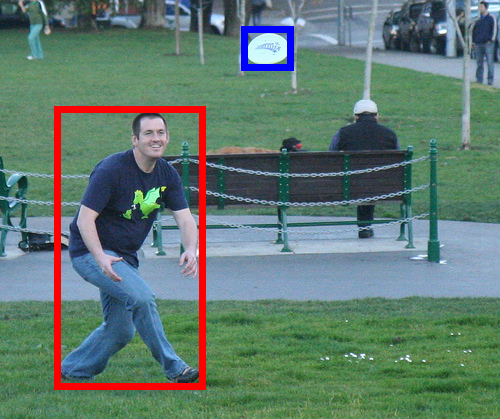}
\includegraphics[width=2.4cm,height=1.6cm]{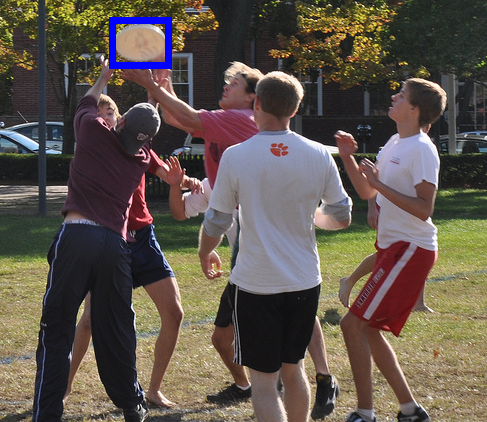}
\includegraphics[width=2.4cm,height=1.6cm]{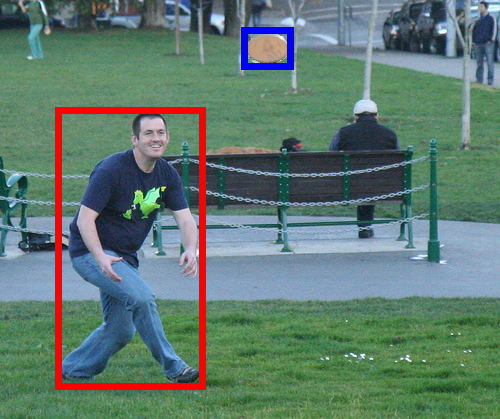}
}
\subfigure{
\includegraphics[width=2.4cm,height=1.6cm]{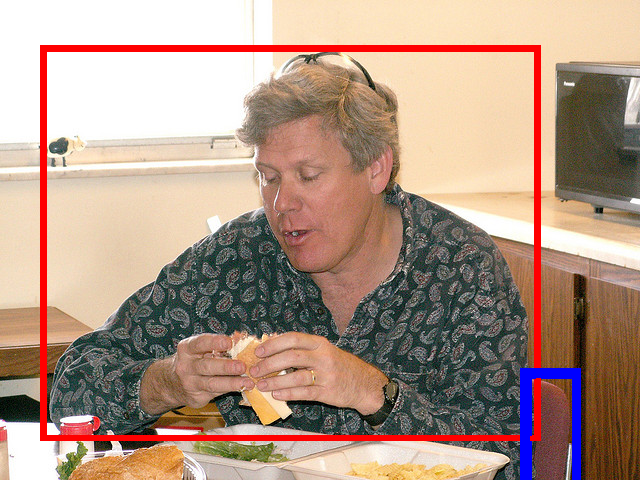}
\includegraphics[width=2.4cm,height=1.6cm]{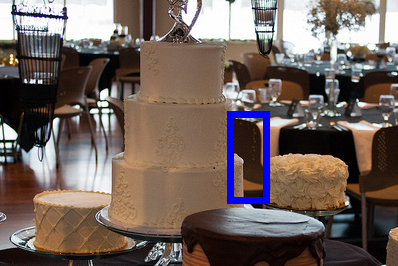}
\includegraphics[width=2.4cm,height=1.6cm]{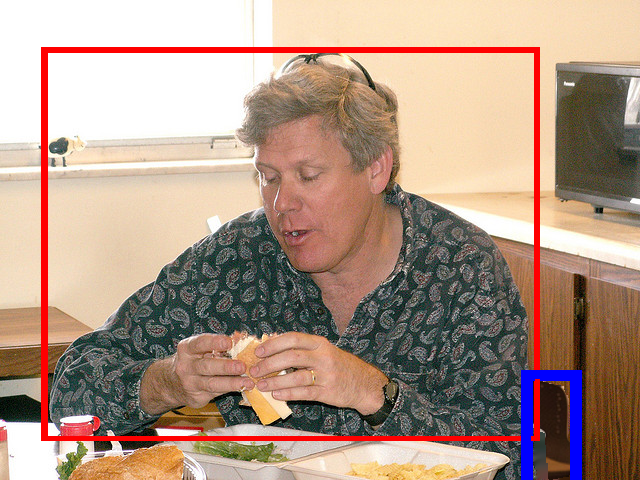}
}
\subfigure{
\includegraphics[width=2.4cm,height=1.6cm]{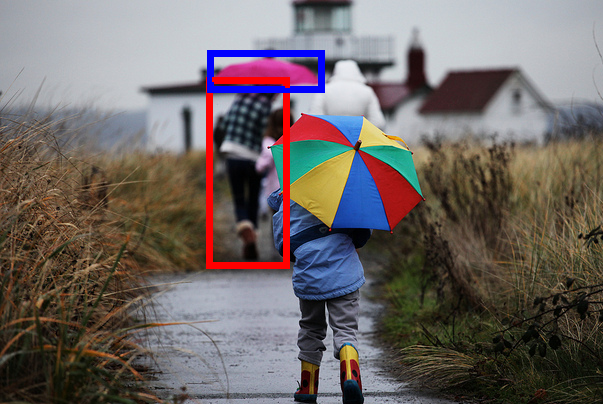}
\includegraphics[width=2.4cm,height=1.6cm]{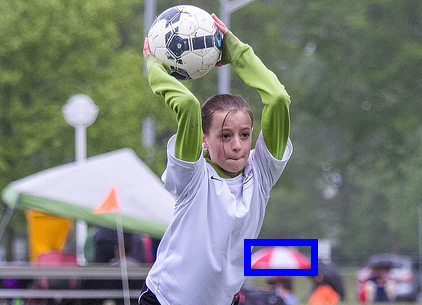}
\includegraphics[width=2.4cm,height=1.6cm]{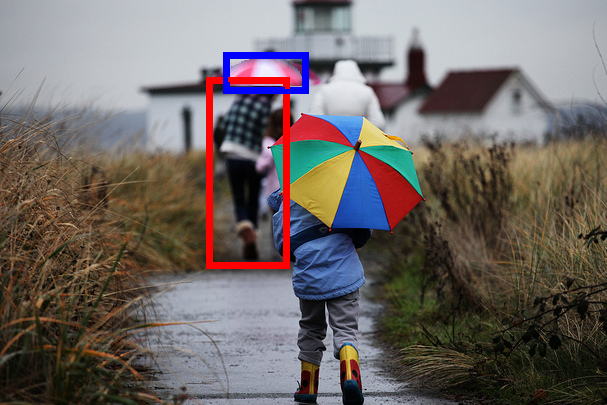}
}
\caption{Left images are original ones. We replace the objects (blue boxes) with instances from the middle images (blue boxes). The rightest images are the augmentation results.}
\label{fig:object exchange}
\end{figure}

\subsubsection{Whether to Replace an Object}
We first judge whether an object can be substituted or not. Some objects are not suitable to be replaced if they interlock with its neighbours too tightly. In this case, adjacent humans or objects are likely to overlap with each other. As a consequence, it is difficult to find a proper replacement to maintain this interaction.

Intuitively, tightly interlocked instances share a long common borderline. Therefore, we develop a metric called \textit{instance interlocking ratio} measuring the interlocking extent between two adjacent instances in the same image.

\begin{figure}[tb!]
\centering
\includegraphics[width=0.3\linewidth]{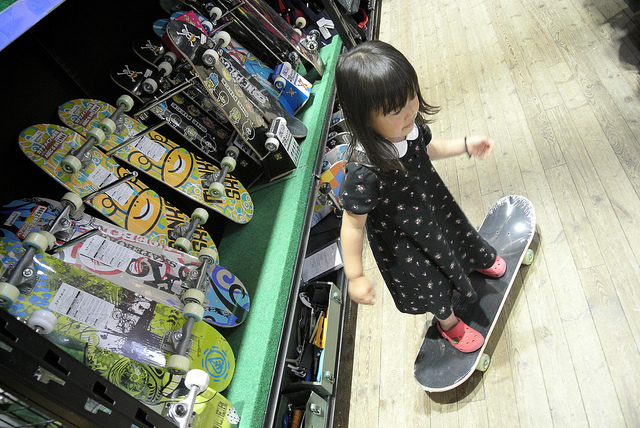}
\includegraphics[width=0.3\linewidth]{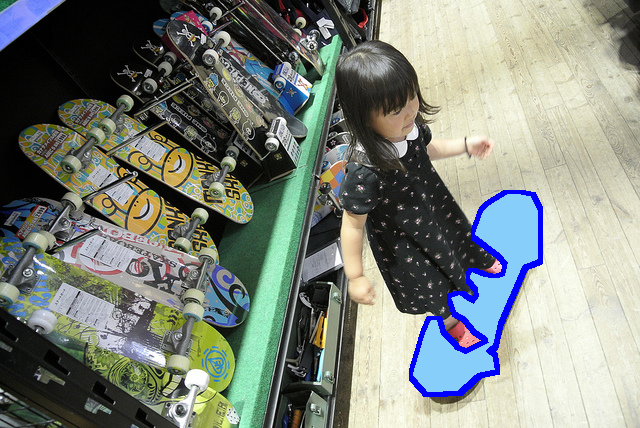}
\includegraphics[width=0.3\linewidth]{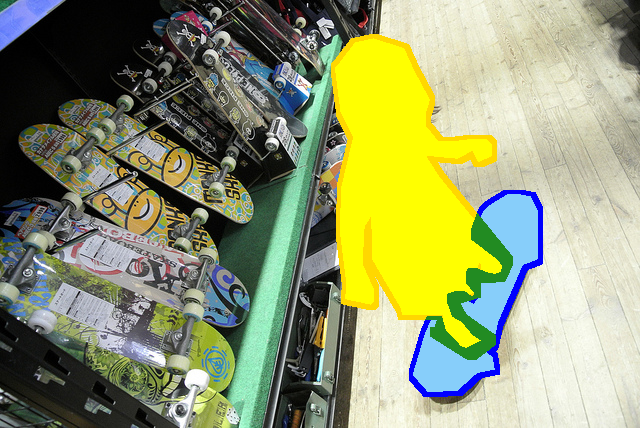}
\caption{In the middle image, light blue region shows the object mask while dark blue denotes the contour. In the right image, for the two instances $O_i,O_j$, $U_{i,j}$ is colored in green and $V_{i,j}$ is composed of the green, dark yellow and dark blue regions.}
\label{fig:contour_mask}
\end{figure}

As shown in Fig.~\ref{fig:contour_mask}, we define $\mathcal{C}_i$ as the contour of instance $O_i$ and $\mathcal{M}_i$ as the mask of this instance. The contour $\mathcal{C}_i$ serves as the outline of the mask with width $w$. For two adjacent instances $O_i,O_j$ in the same image, we define their interlocking area as $U_{i,j}$ and their union contour area as $V_{i,j}$:

\begin{align}
    &U_{i,j} = S(\mathcal{M}_i\cap \mathcal{C}_j)+S(\mathcal{C}_i\cap
    \mathcal{M}_j)
    \label{eq:IA}\\
    &V_{i,j} = S(\mathcal{C}_i\cup \mathcal{C}_j),
    \label{eq:CA}
\end{align}
where $S(\mathcal{A}\cap\mathcal{B})$ denotes the intersection area of $\mathcal{A}$ while $\mathcal{B}$ and $S(\mathcal{A}\cup\mathcal{B})$ denotes the union area of $\mathcal{A}$ and $\mathcal{B}$

Further, the \textit{instance interlocking ratio} between instance $O_i,O_j$ is defined as $r_{i,j}$:

\begin{align}
    r_{i,j}=\frac{U_{i,j}}{V_{i,j}}\in[0,1].
\end{align}

If two adjacent instances have a high interlocking ratio, chances are that they seriously overlap with each other. As a result, neither of them will be replaced. Thus, objects in image $\mathcal{I}$ that can be replaced are selected from the following set:
\begin{align}
    \mathbf{O'} = \left\{O_i |  O_i\in\mathcal{I}, \forall O_j\in\mathcal{I},j\neq i: r_{i,j}<t\right\},
\end{align}
where $t$ is a hyper-parameter as a threshold. We empirically set it to $0.1$.

\subsubsection{Find Objects with Similar States}
Despite the same category, objects show various states including pose variance, shape variance, occlusion variance, etc. Objects to be substituted should be matched with others with similar states. Otherwise, the interaction may be affected. Fortunately, we find that the mask of an object can serve as an indicator of the object state. As the projection of an object on the camera plane given a specific pose, instance mask implicitly encodes the shape and 6D pose of the object. Same category objects may share similar shapes and 6D poses if they have similar masks. What's more, an object's occlusion state can also be reflected from the combination of its own and its neighbours' masks. Thus, we build our object state descriptor based on the object mask.

For object $O_i$ with a $W\times H$ bounding box $\mathcal{X}_i$, we divide $\mathcal{X}_i$ into three parts: object mask $\mathcal{M}_i$, background $\mathcal{B}_i$ and adjacent mask $\mathcal{A}_i$. Based on that, we construct the corresponding \textit{object state matrix} $\textbf{E}_i\in\mathbb{R}^{W\times H}$ for each instance $i$.  Each element in this matrix corresponds with a pixel in the bounding box of instance $i$. The mapping is shown as follows:
\begin{equation}
    \begin{aligned}
        &\textbf{E}_i^{x,y} =
        \begin{cases}
            1 & I_{x,y}\in\mathcal{M}_i\\
            0 & I_{x,y}\in\mathcal{B}_i\\
            -1 & I_{x,y}\in\mathcal{A}_i\\
        \end{cases}\\
        &x\in\{1,\cdots,W\},y\in\{1,\cdots,H\}
    \end{aligned}
    \label{eq:divide region}
\end{equation}
where $I_{x,y}$ denotes the pixel with coordinate $(x,y)$ in the bounding box. This matrix $\textbf{E}_i$ serves as a descriptor of the shape, 6D pose and overlapping condition of instance $O_i$.

With such descriptor, for objects $O_i$ and $O_j$ with state matrix $\textbf{E}_i\in\mathbb{R}^{W\times H}$ and $\textbf{E}_j\in\mathbb{R}^{W'\times H'}$, we define their object state distance $D(i,j)$ as
\begin{equation}
    \begin{aligned}
        &D(i, j)=
        \dfrac{\sum_{x,y}|\mathbf{E_i}-\mathbf{E_j^{’}}|}{W\times H}, \\
        &x\in\{1,2,\cdots,W\},y\in\{1,2,\cdots,H\}
    \end{aligned}
\end{equation}
where $\mathbf{E_j^{’}}$ is the resized matrix of $\mathbf{E_j}$ with same size with $\mathbf{E_i}$.

\begin{figure}[tb!]
\centering
\subfigure[]{
    \includegraphics[height=1.7cm, width=0.4\linewidth]{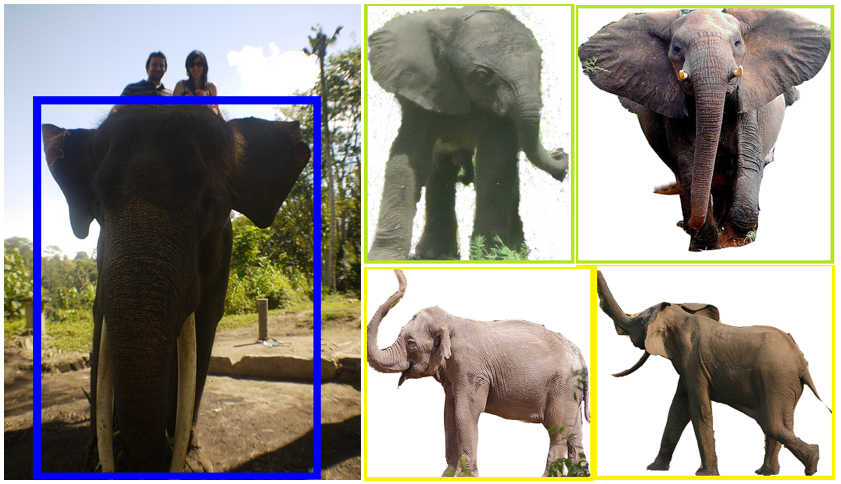}
    \label{fig:similarity(a)}
}
\subfigure[]{
    \includegraphics[height=1.7cm, width=0.5\linewidth]{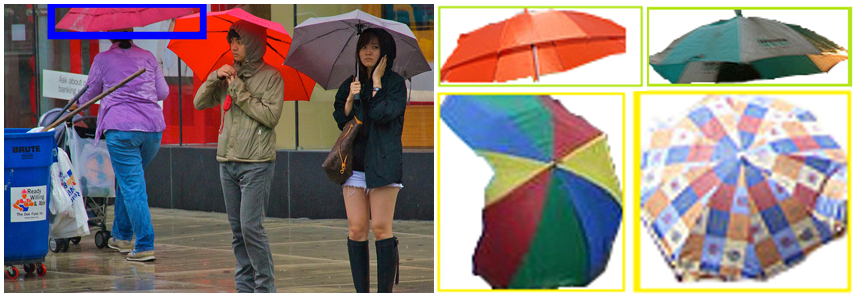}
    \label{fig:similarity(b)}
}
\caption{In (a) and (b), the left is the original image with blue box showing the object. The right above (green box) two images show instances which have high similarity with original object while the right below two (yellow box) in each sub-figure are with low similarity.}
\label{fig:object similarity}
\end{figure}

In the training period, when we process a replaceable object instance $O_i$ in a given image, we randomly select 20 same category objects from other images and calculate their object state distance to $O_i$. Object with the smallest state distance is selected to replace $O_i$. Fig.~\ref{fig:object similarity} shows some positive or negative examples for replacement.

\subsubsection{Object Replacement}
After finding substitution candidate $O_s$ for object $O_i$, we extract both instances from background using instance masks. For datasets without ground-truth segmentation annotations (like HICO-DET), we generate instance masks with Deep Mask \cite{DBLP:journals/corr/PinheiroCD15}. Matting~\cite{he2011global} with alpha channel is adopted to extracted instances so that smoother outlines are acquired. At the same time, we conduct inpainting with Fast Marching~\cite{bertalmio2001navier} to fill the hole of $O_i$ in the background, which ensures the continuous distribution of the raw image. Finally, we resize object $O_s$ to have the same bounding box size as $O_i$ and paste the segmented instance $O_s$ to the original location of object $O_i$.

\begin{figure}[tb!]
\centering
\subfigure[\textit{hold} vs \textit{sit}]{
\includegraphics[width=0.4\linewidth]{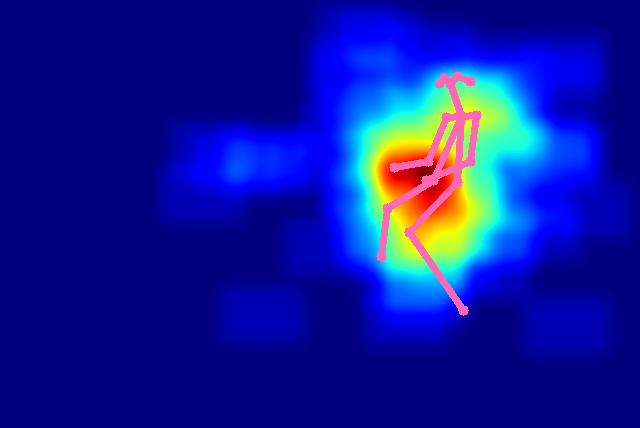}
\includegraphics[width=0.4\linewidth]{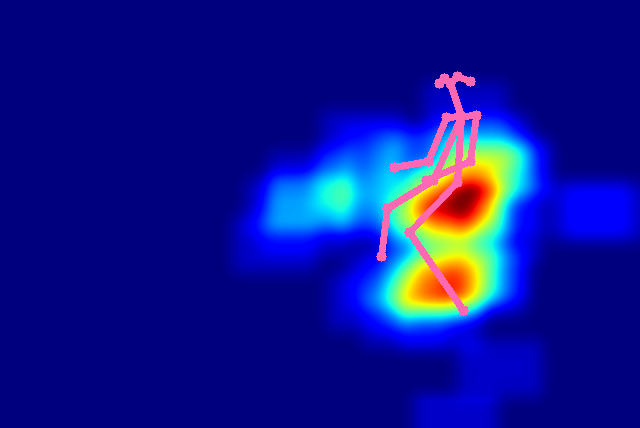}
}
\subfigure[\textit{look} vs \textit{kick}]{
\includegraphics[width=0.4\linewidth]{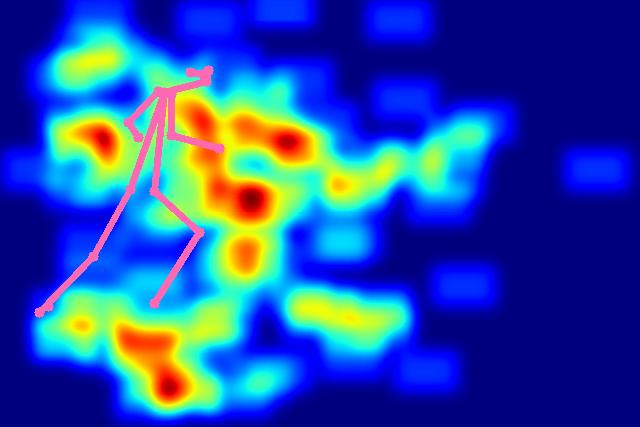}
\includegraphics[width=0.4\linewidth]{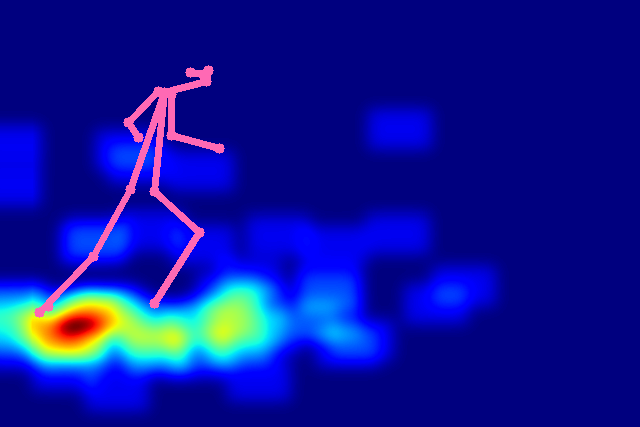}
}

\caption{For same atomic pose, object placement of different interactions has distinct probability distribution. As shown above, objects of \textit{holding} are close to hands, \textit{sitting} close to buttock or legs, \textit{kicking} close to feet, while objects of \textit{looking} extends from eyes and dispersed extensively.}
\label{fig:person map}
\end{figure}

\subsection{Global Spatial Correlation Augmentation}
\label{sec:pose-guided}
In Sec.\ref{sec:object exchange}, the substituted object is pasted at the original position. Although it augments the object appearance, the variance in the image is too slight to cover other unobserved situations. As a supplement, movement with longer distance can effectively improve the performance. In the meantime, such movement should not pose damage to the contextual information.

Therefore, We develop a \textit{pose-guided probability map} to obtain feasible positions of an object. To get the pose data, we follow \cite{li2019transferable} to employ AlphaPose \cite{fang2017rmpe,li2019crowdpose} on each human. The generated pose data $\mathbf{K}$ is in COCO~\cite{lin2014microsoft} format with 17 keypoints of each person.

For each human-object interaction category, the relative spatial correlation between the human and object can be described with a 2-dimension vector $\textbf{v}_{sp}$.

\begin{align}
    \textbf{v}_{sp} = \textbf{c}_o-\textbf{c}_h
\end{align}
where $\textbf{c}_h=(x_{c,h},y_{c,h}),\textbf{c}_o=(x_{c,o},y_{c,o})$ are the torso center of human and bounding box center of object respectively.

We perform normalization to deal with different scales of instances and images. Specifically, torso centers of human poses are set as the origins and torso lengths are normalized to one. Also, the relative spatial position vector $\textbf{v}_{sp}$ is normalized by dividing the torso length. We denote the normalized pose as $\hat{\mathbf{K}}$ and the normalized offset as $\hat{\mathbf{v}}_{sp}$.

To get feasible configurations to augment spatial correlations between human-object pairs, we model the object location $\mathbf{L}$ as a conditional probability distribution \textit{w.r.t} normalized human pose $\hat{\mathbf{K}}$. Considering the proper object location distribution differs across different HOI categories, we learn the conditional distribution for each HOI category separately. Given category $\mathbf{h}$, we model $p(\mathbf{L}|\hat{\mathbf{K}}, \mathbf{h})$ as a mixture of Gaussian distribution. Mathematically, we have
\begin{equation}
    p(\mathbf{L}|\hat{\mathbf{K}},\mathbf{h}) = p(\hat{\mathbf{v}}_{sp}|\mathbf{h}) = \sum_{j=1}^{N_G}\,\omega_{j}\, \mathbb{N}(\hat{\mathbf{v}}_{sp}; \mu_j, \sigma_j),
\end{equation}
where $N_G$ denotes the number of Gaussian distributions, $\omega_{j}$ is the combination weight for the $j$-th component, $\mathbb{N}(\hat{\mathbf{v}}_{sp}; \mu_j, \sigma_j)$ denotes the $j$-th multivariate Gaussian distribution with mean $\mu_{j}$ and covariance $\sigma_j$. Following~\cite{andriluka20142d, fang2018learning}, we set $N_G$ as the number of atomic poses in the dataset, which is 42 in practice. By enforcing the probability distributions independent among each HOI category, we can ensure the object location coherence within each distribution.

We learn the Gaussian mixture distribution $p(\mathbf{L}|\hat{\mathbf{K}}, \mathbf{h})$ efficiently using an EM algorithm, where the E-step estimates the combination weights $\omega$ and M-step updates the Gaussian parameters $\mu$ and $\Sigma$. To simplify the learning process, we utilize K-means clustering to group the pose data in different HOI categories and initialize the parameters as a warm start. Our learned Gaussian Mixture Model (GMM) constitutes the prior knowledge of relative spatial position distribution of the object. The learned mean $\mu_j$ of each Gaussian represents the average of a group of similar 2D poses, which is referred to as atomic pose. Some atomic poses and their corresponding object placement distribution are visualized in Figure~\ref{fig:person map}.

When augmenting an HOI sample in category $\mathbf{h}$ given a human pose $\hat{\mathbf{K}}$, we determine the new relative spatial position vector $\textbf{v}'_{sp}$ by sampling the distribution $p(\mathbf{L}|\hat{\mathbf{K}}, \mathbf{h})$. The augmentation process was illustrated in Fig.~\ref{fig:overview}. Objects are more likely to be placed in a relative spatial position with more prior samples of current interaction type, where they share human poses of the same cluster. With our pose-guided probability map, we are able to augment the spatial correlations between humans and objects in an effective manner.

\section{Experiments}
In this section, we first describe the datasets and metrics. We then introduce the base models on which DecAug is performed, including other implementation details. Next, improvements brought by our method is revealed. Analysis shows that our methods alleviate the long-tail issue. Detailed ablation studies are also conducted.

\subsection{Dataset and Metric}
\paragraph{Dataset}
We evaluate our methods on two mainstream benchmarks: \textbf{V-COCO}~\cite{gupta2015visual} and \textbf{HICO-DET}~\cite{chao2018learning}. \textbf{V-COCO} is a subset of COCO dataset~\cite{lin2014microsoft} annotated with HOI labels. It includes 10,346 images (2,533 for training, 2,867 for validating and 4,946 for testing) and 16,199 human instances. Each person is annotated with 29 action types, 5 of which have no object. The objects are split into two types: \textit{object} and \textit{instrument}. \textbf{HICO-DET} is a subsect of HICO~\cite{chao2015hico} dataset annotated with bounding boxes. It contains 47,776 images (38,118 for training and 9,658 for testing), 600 HOI categories over 80 object types and 117 verbs.

\paragraph{Metric}
We apply the mainstream metric for HOI detection: role mean average precision (role mAP).
A prediction is true positive only when
1) HOI classification is correct, and
2) both the IoUs between the predicted bounding boxes of human and object v.s. the ground truth $>0.5$.

\subsection{Implementation Details}
\paragraph{Models}

We apply DecAug to the following two representative HOI detection models: iCAN~\cite{gao2018ican} and Transferable Interactiveness Network (TIN)~\cite{li2019transferable}.
 Same object proposals are applied so that we can ensure the performance gain comes from interaction recognition instead of object detection.
 Baseline results are those reported in their original papers.

\paragraph{Hyper-parameters}
We adopt stochastic gradient descent in training. All hyper-parameters strictly follow the original setting of our baseline models including iteration number, learning rate, weight decay, backbones and so on.
\paragraph{Augmentation Pipeline}
During training, the proposed local and global augmentation strategies are incorporated simultaneously since they are complimentary. Each input image will be augmented with a probability of 0.5.

\begin{table}[tb!]
\begin{center}
\caption{\textbf{Results on V-COCO}: Original models' results come from their papers.}
\label{tab:vcoco res}
\begin{tabular}{lcc}
\toprule
\textbf{Model} & \textbf{DecAug} & $\textbf{mAP}_{role}$\\
\midrule
iCAN~\cite{gao2018ican} & & 44.7\\
iCAN & \ding{51} & \textbf{48.0}\\
\textit{Improvement} & &  \textbf{3.3}$\mathbf{\uparrow}$\\
\midrule
TIN ($\textbf{R}\textbf{P}_D\textbf{C}_D$)~\cite{li2019transferable} & & 47.8\\
TIN ($\textbf{R}\textbf{P}_D\textbf{C}_D$) & \ding{51} & \textbf{50.4}\\
\textit{Improvement} & &  \textbf{2.6}$\mathbf{\uparrow}$ \\
\bottomrule
\end{tabular}
\end{center}
\end{table}

\begin{table*}[tb!]
\begin{center}
\caption{\textbf{Results on HICO-DET}: Original models' results come from their papers.}
\label{tab:hico res}
\begin{tabular}{lccccccc}
\toprule
\multirow{2}{*}{\textbf{Model}} & \multirow{2}{*}{\textbf{DecAug}} & \multicolumn{3}{c}{\textbf{Default}} & \multicolumn{3}{c}{\textbf{Known Object}}\\
& & \textbf{Full} & \textbf{Rare} & \textbf{Non-Rare} & \textbf{Full} & \textbf{Rare} & \textbf{Non-Rare} \\
\midrule
iCAN~\cite{gao2018ican} & & 14.84 & 10.45 & 16.15 & 16.26 & 11.33 & 17.73\\
iCAN &\ding{51} & \textbf{16.39} & \textbf{12.23} & \textbf{17.63} & \textbf{17.85} & \textbf{13.68} & \textbf{19.10}\\
\textit{Improvement} & & \textbf{1.55}$\mathbf{\uparrow}$ &\textbf{1.78}$\mathbf{\uparrow}$ &\textbf{1.48}$\mathbf{\uparrow}$ &\textbf{1.59}$\mathbf{\uparrow}$ &\textbf{2.35}$\mathbf{\uparrow}$ &\textbf{1.37}$\mathbf{\uparrow}$\\
\midrule
TIN ($\textbf{R}\textbf{P}_D\textbf{C}_D$)~\cite{li2019transferable} & & 17.03 & 13.42 & 18.11 &19.17& 15.51& 20.26 \\
TIN ($\textbf{R}\textbf{P}_D\textbf{C}_D$) &\ding{51} & \textbf{18.38} & \textbf{14.99} & \textbf{19.39} & \textbf{20.50} & \textbf{16.93} & \textbf{21.57}\\
\textit{Improvement} & &\textbf{1.35}$\mathbf{\uparrow}$ &\textbf{1.57}$\mathbf{\uparrow}$ &\textbf{1.28}$\mathbf{\uparrow}$ &\textbf{1.33}$\mathbf{\uparrow}$ &\textbf{1.42}$\mathbf{\uparrow}$ &\textbf{1.31}$\mathbf{\uparrow}$ \\
\bottomrule
\end{tabular}
\end{center}
\end{table*}

\begin{figure*}[tb!]
\centering

\subfigure[iCAN with DecAug]{
\includegraphics[width=2.5cm,height=1.8cm]{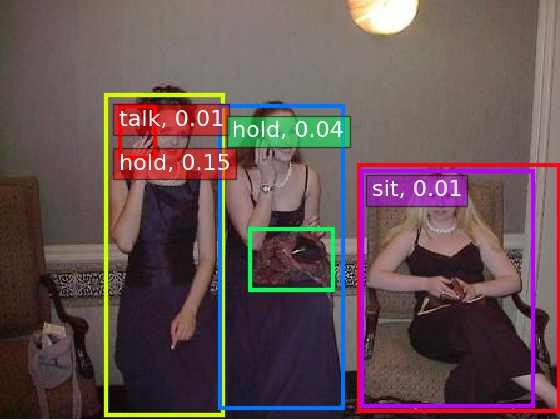}
\includegraphics[width=2.5cm,height=1.8cm]{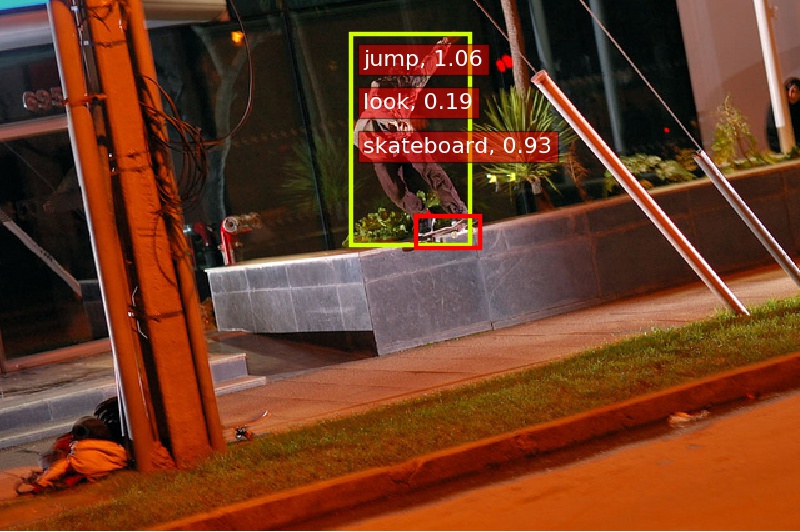}
\includegraphics[height=1.8cm]{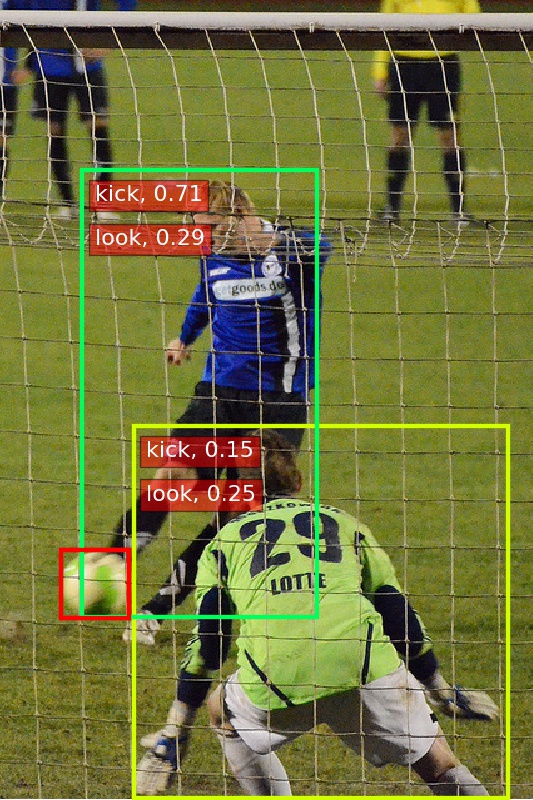}
\includegraphics[width=2.5cm,height=1.8cm]{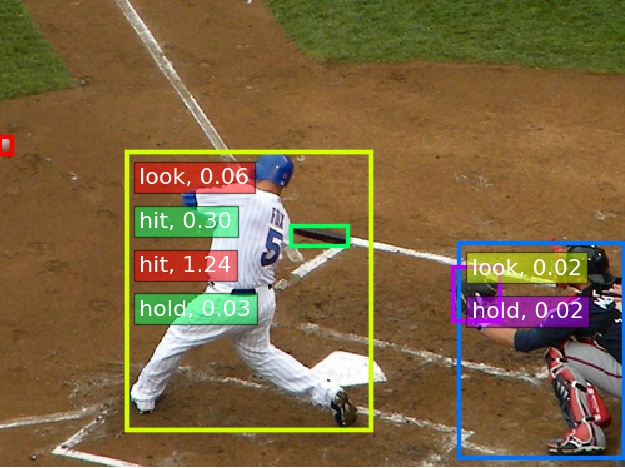}
\includegraphics[height=1.8cm]{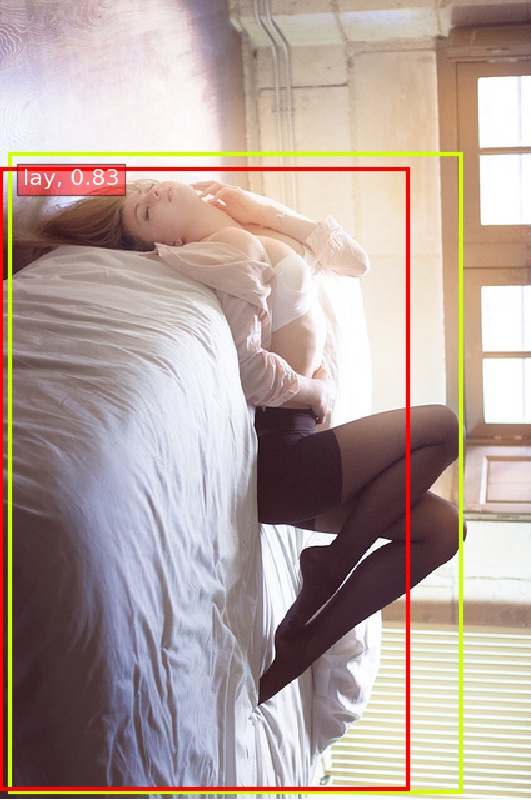}
\includegraphics[width=2.5cm,height=1.8cm]{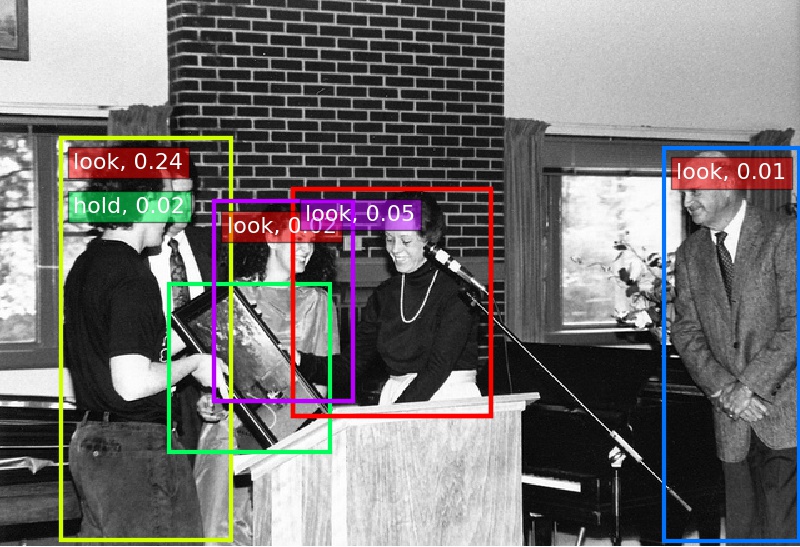}
\includegraphics[width=2.5cm,height=1.8cm]{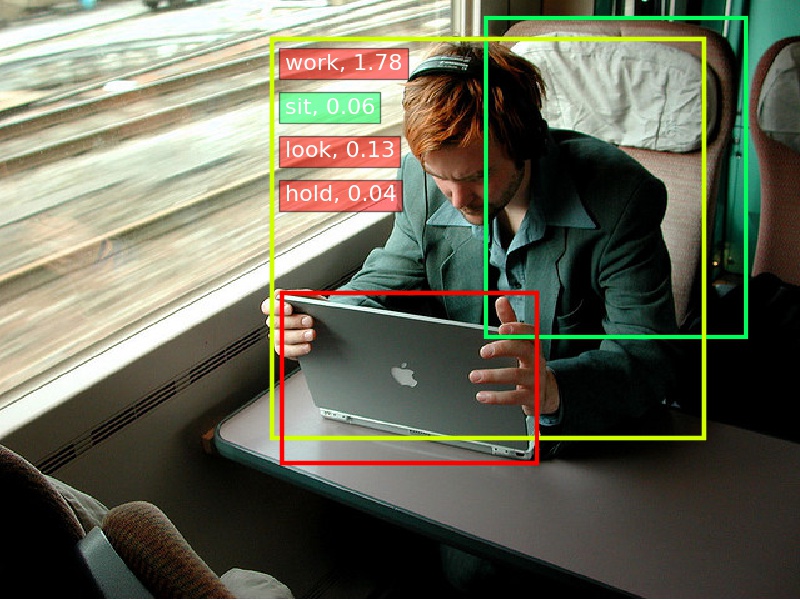}
\includegraphics[height=1.8cm]{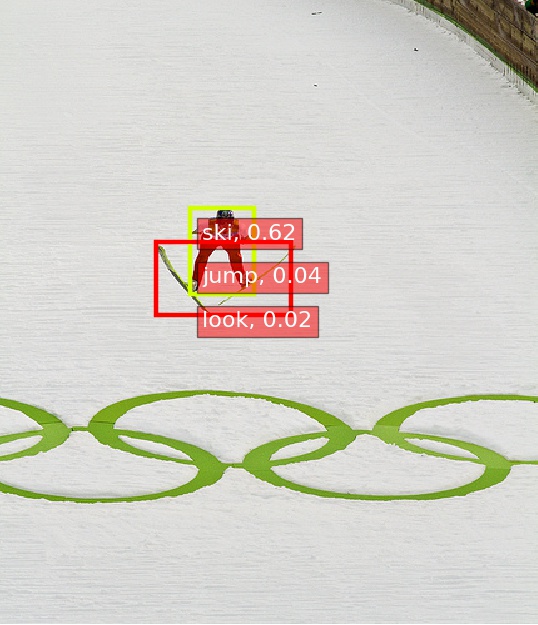}
}

\vfill

\subfigure[iCAN w/o DecAug]{
\includegraphics[width=2.5cm,height=1.8cm]{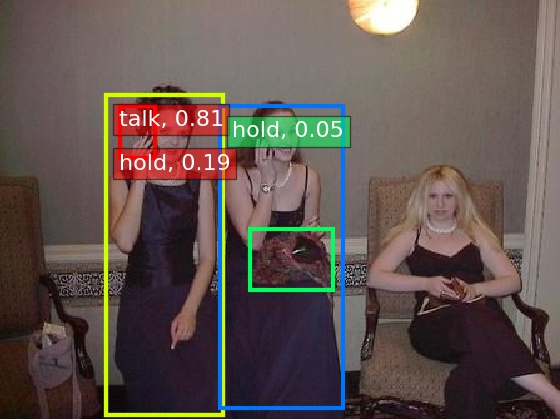}
\includegraphics[width=2.5cm,height=1.8cm]{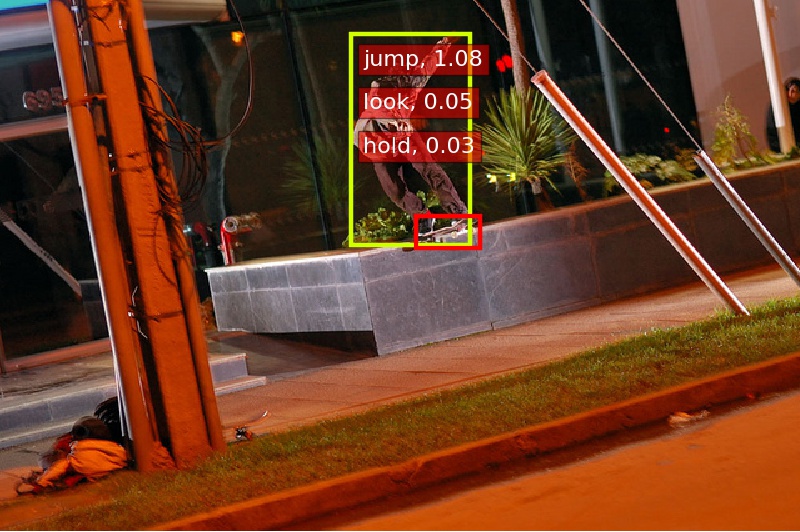}
\includegraphics[height=1.8cm]{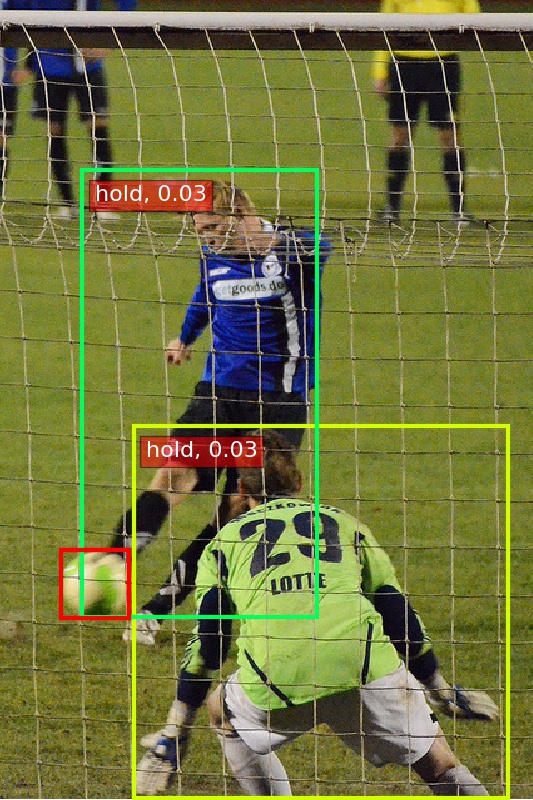}
\includegraphics[width=2.5cm,height=1.8cm]{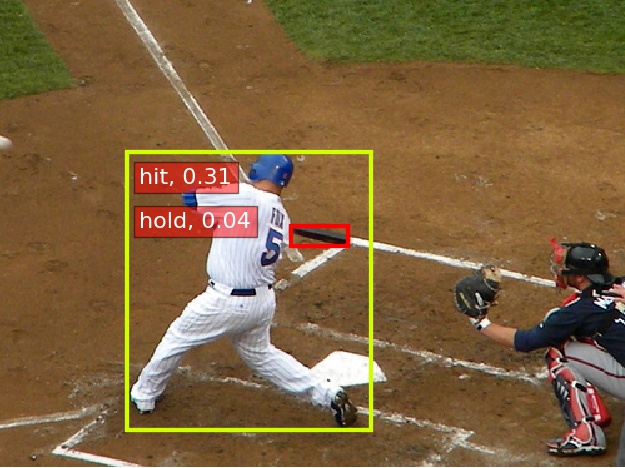}
\includegraphics[height=1.8cm]{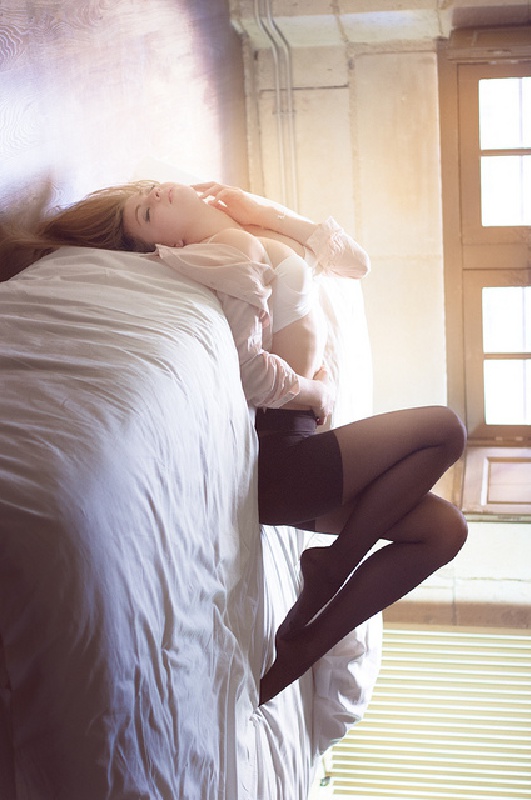}
\includegraphics[width=2.5cm,height=1.8cm]{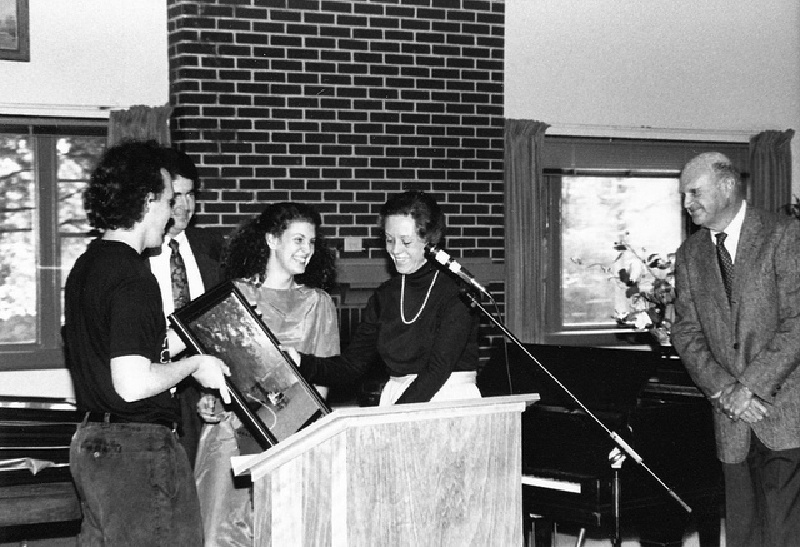}
\includegraphics[width=2.5cm,height=1.8cm]{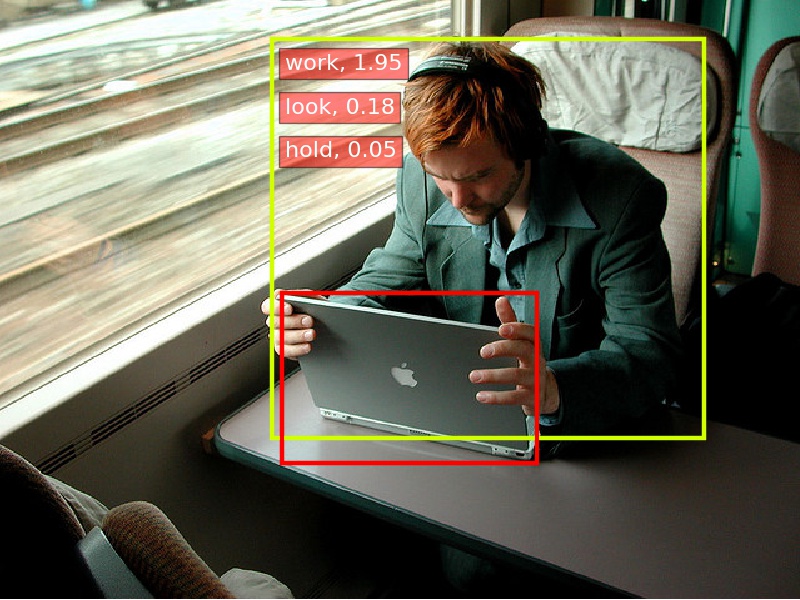}
\includegraphics[height=1.8cm]{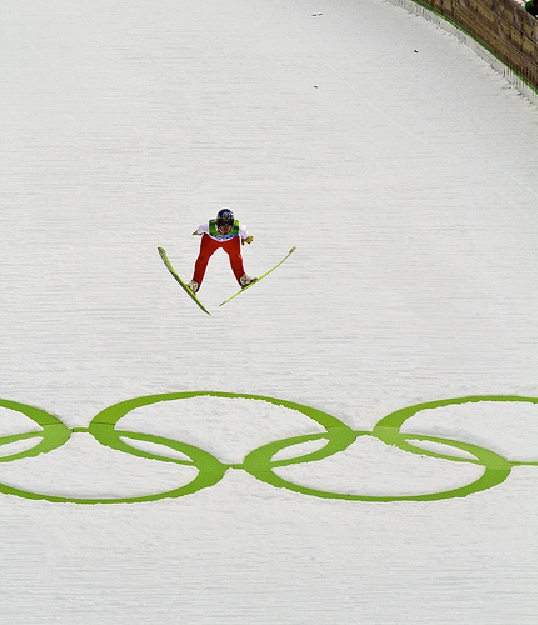}
}
\caption{HOI detection results of iCAN trained with (top) and w/o (bottom) DecAug. DecAug brings more accurate detection.}
\label{fig:vis ican}
\end{figure*}

\subsection{Results and Comparison}
\label{sec:exp}

The HOI detection results are evaluated by following the detailed metrics defined by each specific dataset. Results of all the experiments verify the effectiveness and generality of the proposed DecAug.

For \textbf{V-COCO}, we evaluate $mAP_{role}$ in Tab.~\ref{tab:vcoco res}. 
We can see that substantial improvements (3.3 mAP) are achieved by applying DecAug.

For \textbf{HICO-DET}, we evaluate $mAP_{role}$ of Full (600 HOIs), Rare (138 HOIs), Non-Rare (462 HOIs) interactions of two different settings: Default and Known Object. Results are shown in Tab.~\ref{tab:hico res}. 
Unsurprisingly, notable performance gain is also achieved (1.6 mAP), indicating the effectiveness of our methods on large datasets without ground-truth segmentation or keypoints.

In Fig~\ref{fig:vis ican}, we show some visualized results trained with and w/o DecAug. We can see examples that our DecAug compensates for some ignorance and corrects some detection mistakes, as it makes full use of the information within the whole dataset.

\subsection{Analysis}
\begin{figure}[t]
\centering

\subfigure[]{
\includegraphics[width=3.8cm,height=2.4cm]{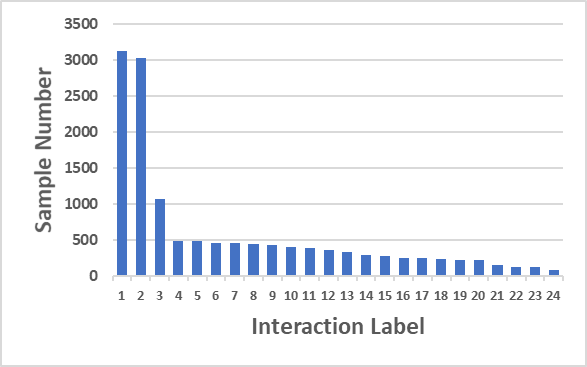}
\label{fig:long-tail(a)}
}
\subfigure[]{
\includegraphics[width=3.8cm,height=2.4cm]{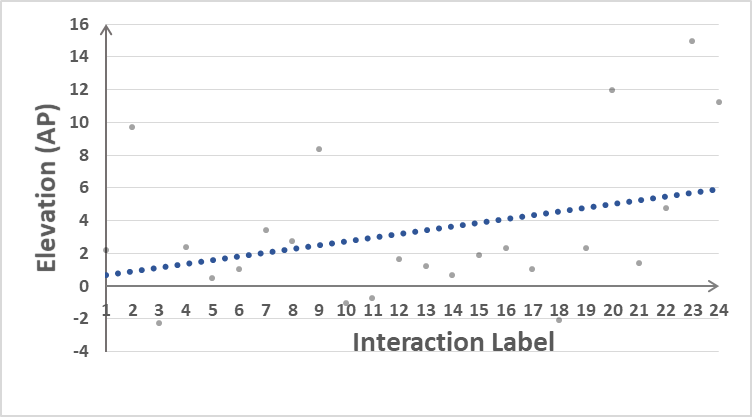}
\label{fig:long-tail(b)}
}
\caption{(a) shows the training sample number of each interaction category in V-COCO dataset. Interaction names are ignored for clarity. Grey points in (b) show the ${AP}_{role}$ improvement of each interaction category (corresponding with (a)). The blue dotted line in (b) reveals the fitted trend line of ${AP}_{role}$ improvement. We can see that the elevation increases as the sample number decreases.}
\label{fig:long-tail}
\end{figure}

\paragraph{Long-tail Issue}
is a pervasive problem in HOI datasets. In Fig.~\ref{fig:long-tail(a)}, we plot the number of samples from each interaction categories in V-COCO dataset. Severe data imbalance could be observed.  Fig.~\ref{fig:long-tail(b)} then shows the effectiveness of DecAug, from which we can clearly see that more remarkable improvement could be made for interaction categories with fewer training samples. This is because DecAug could make full use of favourable information (e.g. object appearance, spatial locations) across the whole dataset.

\paragraph{Transferability and generality}


It's noteworthy that besides making the best of the prior knowledge in the current dataset, our DecAug is also capable of capturing information from other datasets. This could be achieved since both our object state descriptor and pose-guided probability map can be transferred across datasets. For local object appearance augmentation, annotated instances from other large datasets (e.g. COCO \cite{lin2014microsoft}) may also serve as candidates for replacement.
For global spatial correlation augmentation, the Gaussian mixture model can be constructed based on human-object pairs with similar interactions from the mixed dataset.

In Tab.~\ref{tab:transfer res}, we show the results of applying DecAug on HICO-DET dataset with information transferred from the mixture data of COCO and V-COCO dataset. This well demonstrates the generality of DecAug (such results are not listed in the main table to avoid unfair comparison).

\begin{table*}[tb!]
\begin{center}
\caption{\textbf{Dataset Transferability}: Last two lines show the results on HICO-DET with extra information transferred from the COCO or V-COCO dataset.}
\label{tab:transfer res}
\begin{tabular}{ccccccccc}
\toprule
\multirow{2}{*}{\textbf{Model}} & \multirow{2}{*}{\textbf{DecAug}} &\multirow{2}{*}{\textbf{Transfer}} & \multicolumn{3}{c}{\textbf{Default}} & \multicolumn{3}{c}{\textbf{Known Object}}\\
& & &\textbf{Full} & \textbf{Rare} & \textbf{Non-Rare} & \textbf{Full} & \textbf{Rare} & \textbf{Non-Rare} \\
\midrule
\multirow{4}{*}{iCAN} & & &14.84 & 10.45 & 16.15 & 16.26 & 11.33 & 17.73\\
 &\ding{51} & \ding{55} &16.39 & 12.23 & 17.63 & 17.85 & 13.68 & 19.10\\
 &\ding{51} &objects &\textbf{16.65} & \textbf{12.28} & \textbf{17.96} & \textbf{18.09} & 13.49& \textbf{19.47}\\
 &\ding{51} &spatial &\textbf{16.56} & \textbf{12.32} & \textbf{17.83} & \textbf{18.15} & \textbf{13.70}& \textbf{19.48}\\
\bottomrule
\end{tabular}
\end{center}
\end{table*}

\paragraph{Training Efficiency}

As a data augmentation method, DecAug can be embedded into various existing models conveniently with negligible offline data preprocessing.
During training, it could generate augmented samples online without burdening GPUs. As shown in Tab.~\ref{tab:ablation step}, when applying multi-threads data loader, the training efficiency almost remains unaffected.

\subsection{Ablation Study}

In this part, the impact of
1) local object appearance augmentation (LOA), and 2) global spatial correlation augmentation (GSC)
in DecAug is separately analyzed.
The results are shown in Tab.~\ref{tab:ablation step}.
We can see that both strategies contribute notably to the final performance. Next, we evaluate the effectiveness of some key techniques in each strategy.

\begin{table}[htb!]
\begin{center}
\caption{\textbf{Ablation Study by Removing Either Component}: LOA denotes local object appearance augmentation and GSC denotes global spatial correlation augmentation.}
\label{tab:ablation step}
\begin{tabular}{ccccc}
\toprule
\textbf{Model} & \textbf{LOA} & \textbf{GSC} &\textbf{Train Rate ($s/it$)}& $\textbf{mAP}_{role}$ \\
\midrule
\multirow{4}{*}{iCAN} & & &0.183 &44.7\\
& \ding{51}& &0.191 &46.8\\
& & \ding{51}& 0.190 &47.2\\
&\ding{51}& \ding{51} &0.193 &48.0\\
\bottomrule
\end{tabular}
\end{center}
\end{table}

\paragraph{Local Object Appearance Augmentation}

Here we evaluate the two key components in LOA, \textit{instance interlocking ratio (IIR)} and \textit{object state matrix (OSM)}, by replacing them with other possible metrics.
For IIR, we try other two possible choices: simply replacing all objects (replace all) and applying bbox IoU between neighbours as the metric (bbox IoU). For OSM, we also select other four alternatives: random selection, chamfer distance, instance mask distance and $l_2$ distance of the image inside a bounding box.
In Tab.~\ref{tab:ablation exchange}, results show apparent degradation using other metrics, verifying the significance of our proposed metric.

\paragraph{Global Spatial Correlation Augmentation}
Global spatial correlation augmentation can greatly increase the data diversity without harming the context. We exhibit its value by comparing our results with the other two possible choices: random placement and appearance consistent metric \textit{heatmap} in \cite{fang2019instaboost}. Tab.~\ref{tab:ablation pose} reveals that performance drops notably
with the other alternatives, further proving the power of our pose-guided method.

\begin{table}[htb!]
\caption{\textbf{Ablation Study of Object Appearance and Spatial Correlation Augmentation}}
\centering
\subtable[\textbf{Local Object Appearance Augmentation Ablation Study}: Apply other alternative interchangeability metrics or object similarity metrics. IIR and OSM denote \textit{instance interlocking ratio} and \textit{object state matrix} respectively]{
       \label{tab:ablation exchange}
       \begin{tabular}{llc}
       \toprule
       \textbf{Interchangeability} & \textbf{Similarity} & $\textbf{mAP}_{role}$\\
       \midrule
        IIR & random & 46.6\\
        IIR & chamfer distance & 47.2\\
        IIR & mask distance & 47.5\\
        IIR & bbox distance & 47.1\\
        replace all & OSM & 47.1\\
        bbox IoU & OSM & 47.5\\
        IIR & OSM & 48.0 \\
        \bottomrule
      \end{tabular}
}
\qquad
\subtable[\textbf{Global Spatial Correlation Augmentation}: we compare three placement metrics: random, heatmap~\cite{fang2019instaboost} and our pose-guided GMM.]{
        \label{tab:ablation pose}
       \begin{tabular}{p{5cm}c}
       \toprule
       \textbf{Object Placement Approach} & $\textbf{mAP}_{role}$\\
       \midrule
       random & 43.6\\
       heatmap & 45.3\\
       pose-guided GMM & 48.0\\
        \bottomrule
      \end{tabular}
}
\end{table}

\section{Conclusion}
In this paper, we propose a novel data augmentation method, \textbf{DecAug}, for HOI detection, which mainly includes two components: local object appearance augmentation and global spatial correlation augmentation.
With negligible cost, our method can be easily combined with various existing models to further improve their performance.
DecAug has good generalizability, which could utilize information transferred from other datasets, and it helps address the long-tail problem.
We hope our DecAug gives a new insight into the data augmentation of visual relationship detection.

\small{
\bibliographystyle{ieee}
\bibliography{egbib}
}

\newpage

\begin{figure*}[htb!]
\centering
\caption{We show some augmented images as below. In each row, the first image is the original input (red box: human, blue box: object). The second image is the result of local object appearance augmentation. Some objects are not changed because they do not meet our requirement (instance interlocking ratio) in the paper. The third and forth images show the \textit{pose-guide probability map} and the result of global spatial correlation augmentation.}
\subfigure[sit on the chair]{
\includegraphics[width=0.2\linewidth]{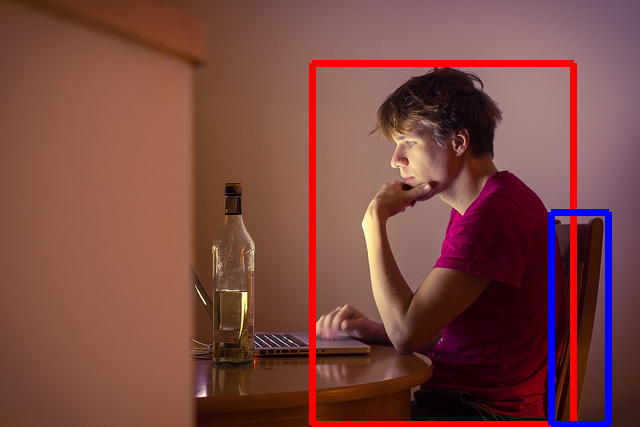}
\includegraphics[width=0.2\linewidth]{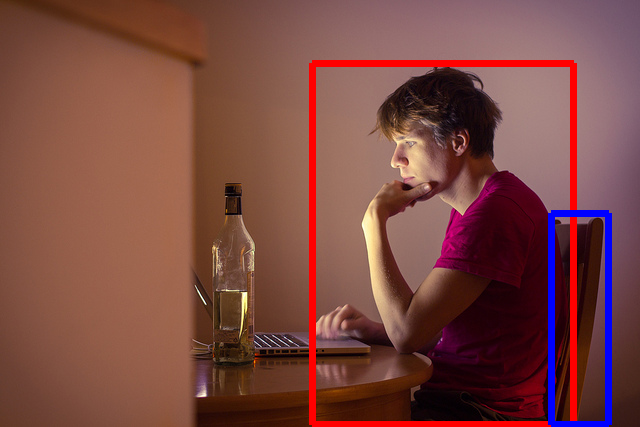}
\includegraphics[width=0.2\linewidth]{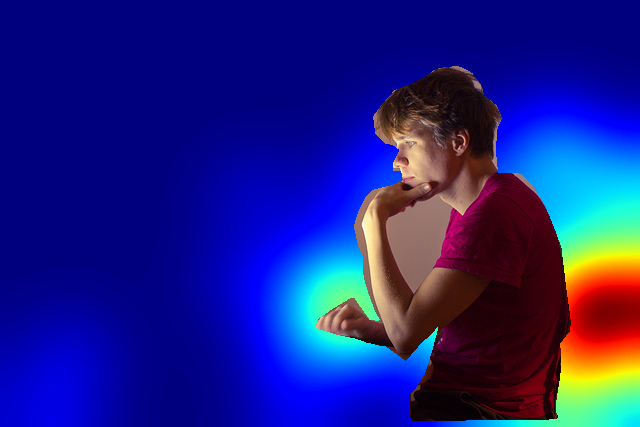}
\includegraphics[width=0.2\linewidth]{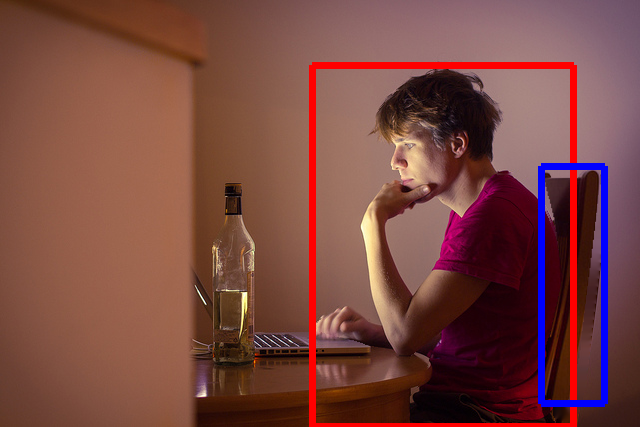}
}
\subfigure[surf with the board]{
\includegraphics[width=0.2\linewidth]{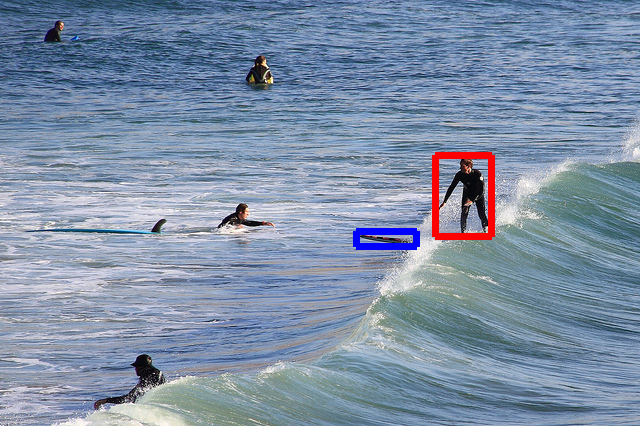}
\includegraphics[width=0.2\linewidth]{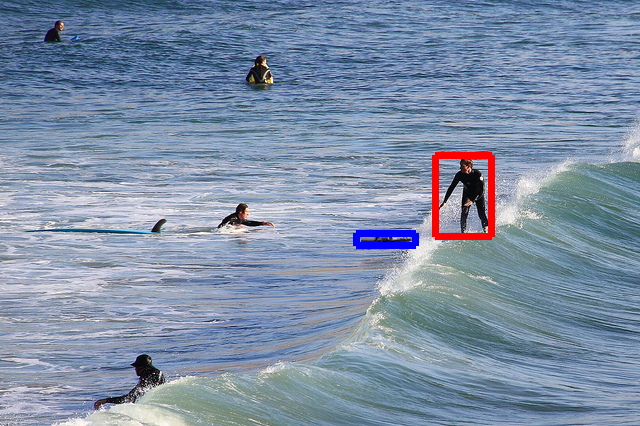}
\includegraphics[width=0.2\linewidth]{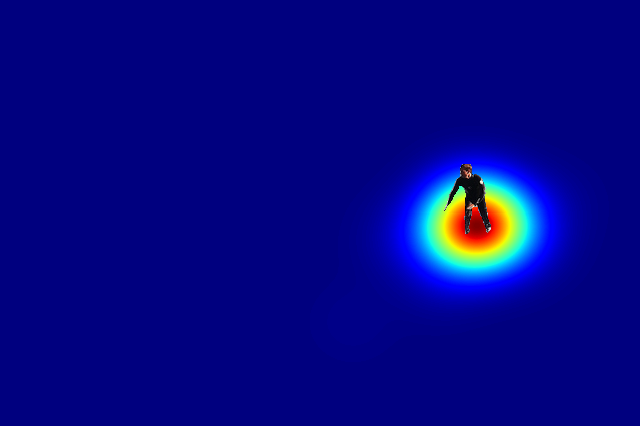}
\includegraphics[width=0.2\linewidth]{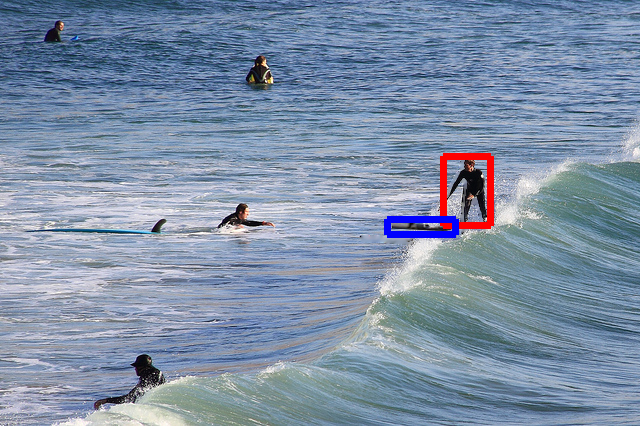}
}
\subfigure[snowboard with the board]{
\includegraphics[width=0.2\linewidth]{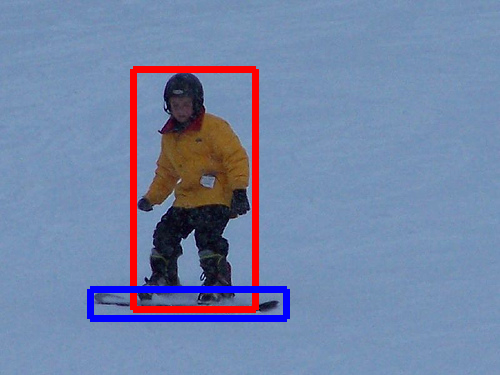}
\includegraphics[width=0.2\linewidth]{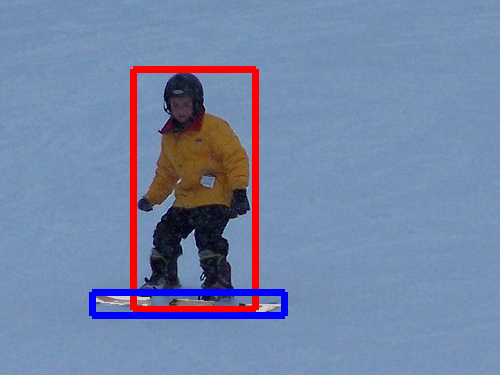}
\includegraphics[width=0.2\linewidth]{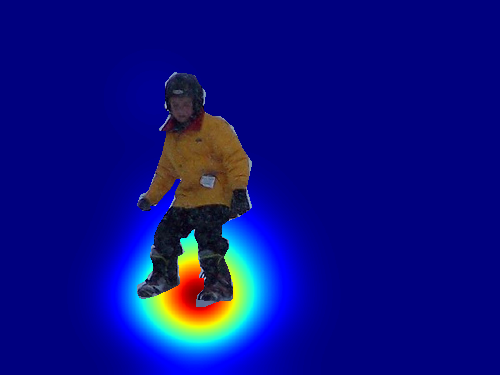}
\includegraphics[width=0.2\linewidth]{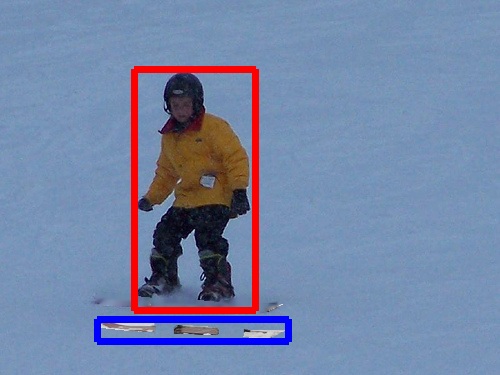}
}
\subfigure[ride the horse]{
\includegraphics[width=0.2\linewidth]{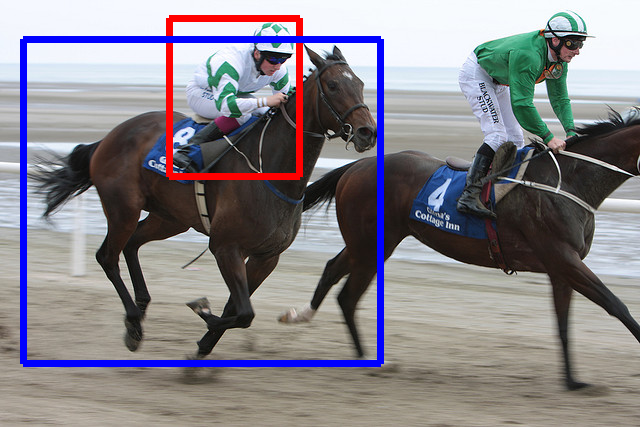}
\includegraphics[width=0.2\linewidth]{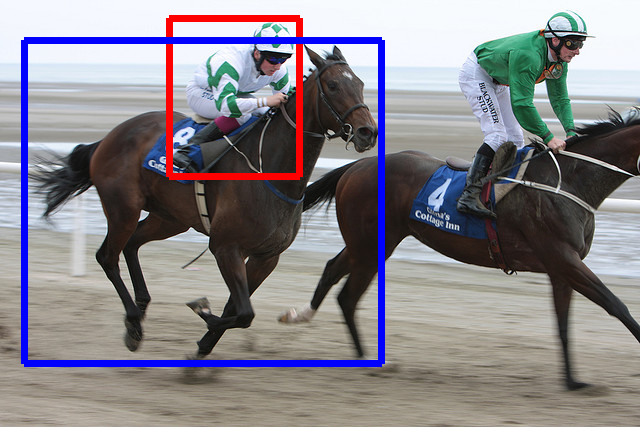}
\includegraphics[width=0.2\linewidth]{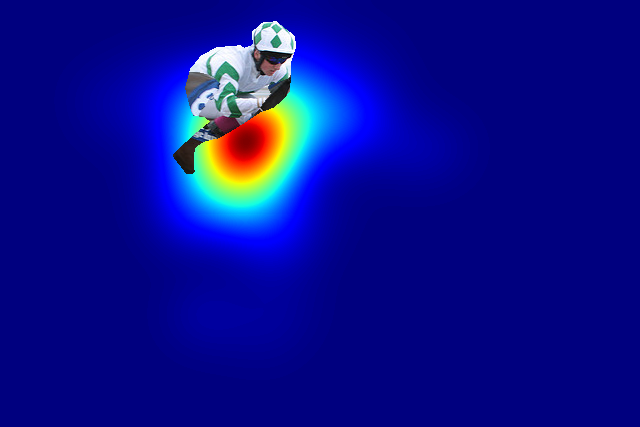}
\includegraphics[width=0.2\linewidth]{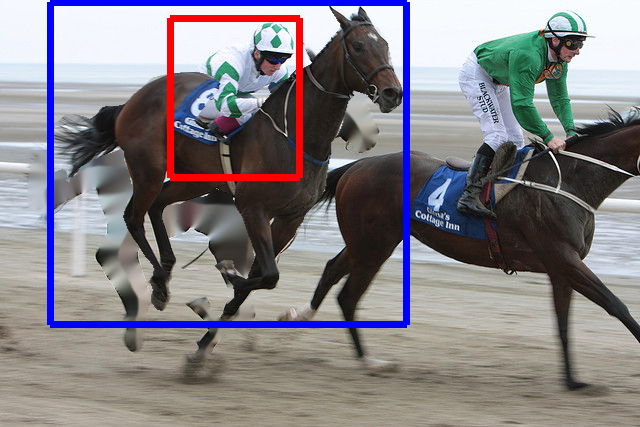}
}
\end{figure*}

\begin{figure*}[tb!]
\centering
\subfigure[surf with the board]{
\includegraphics[width=0.2\linewidth]{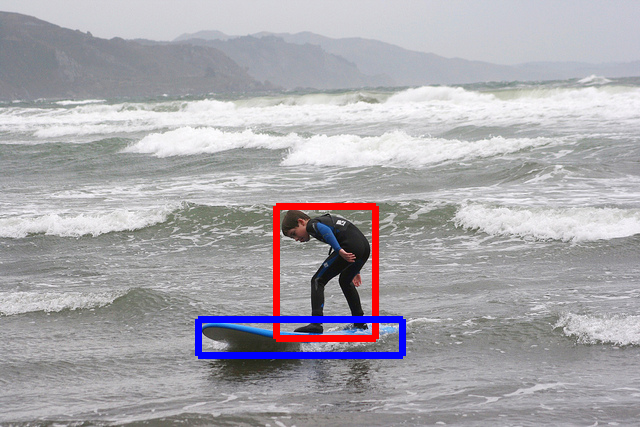}
\includegraphics[width=0.2\linewidth]{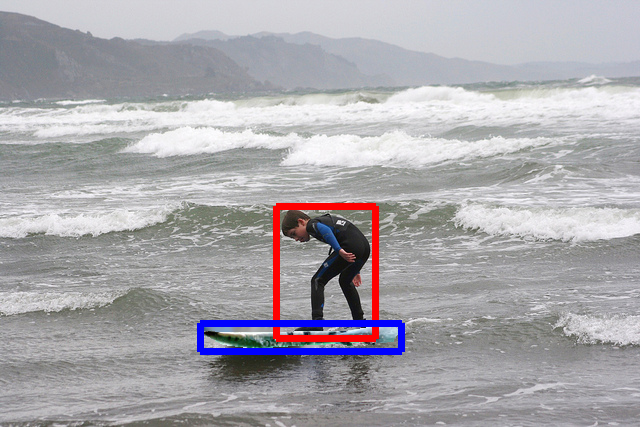}
\includegraphics[width=0.2\linewidth]{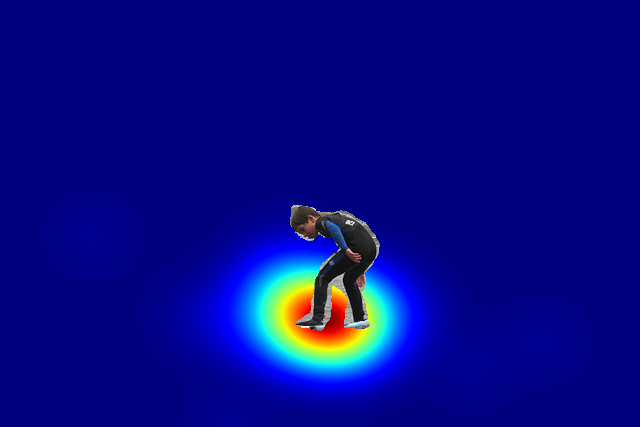}
\includegraphics[width=0.2\linewidth]{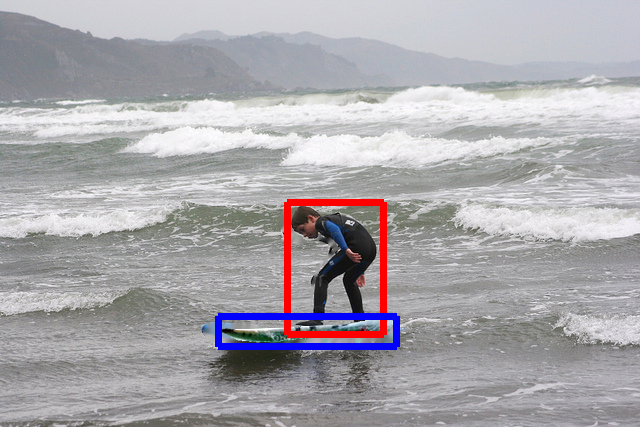}
}
\subfigure[sit on the chair]{
\includegraphics[width=0.2\linewidth]{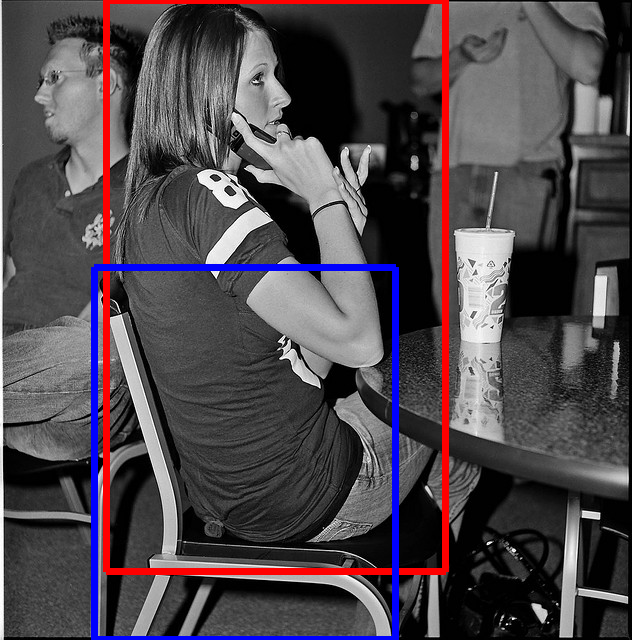}
\includegraphics[width=0.2\linewidth]{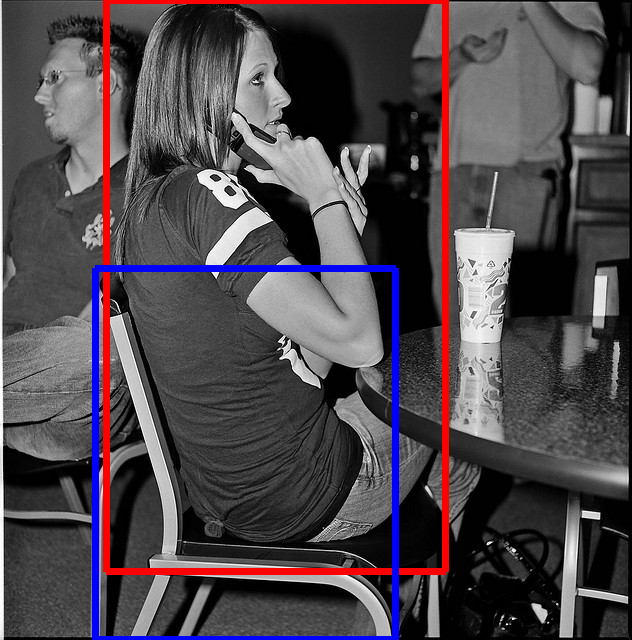}
\includegraphics[width=0.2\linewidth]{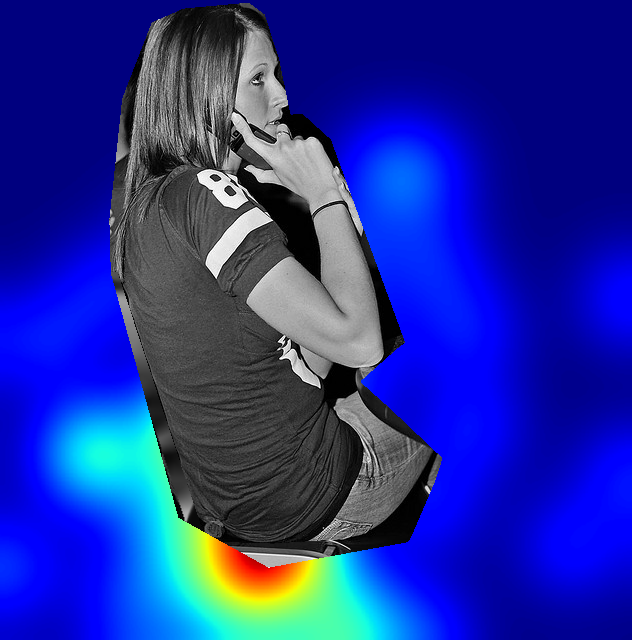}
\includegraphics[width=0.2\linewidth]{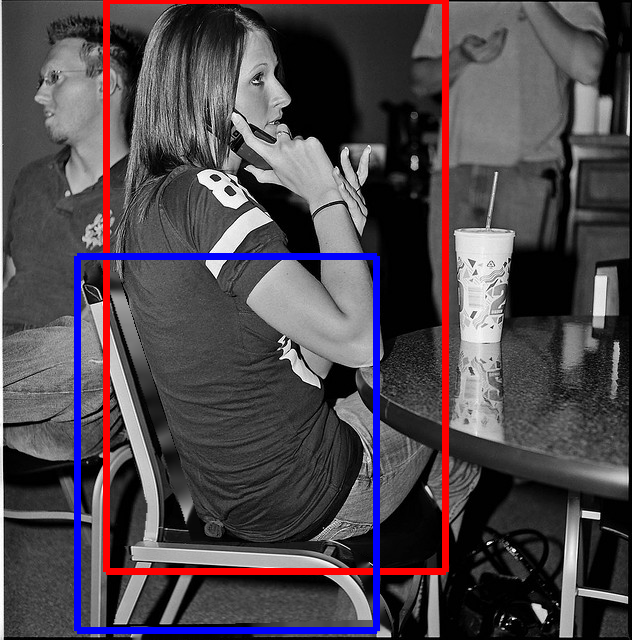}
}
\subfigure[hit/look the ball]{
\includegraphics[width=0.2\linewidth]{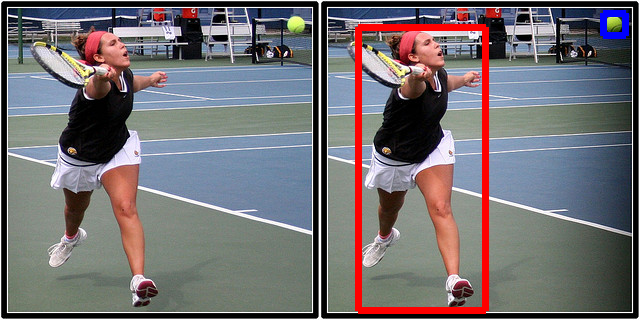}
\includegraphics[width=0.2\linewidth]{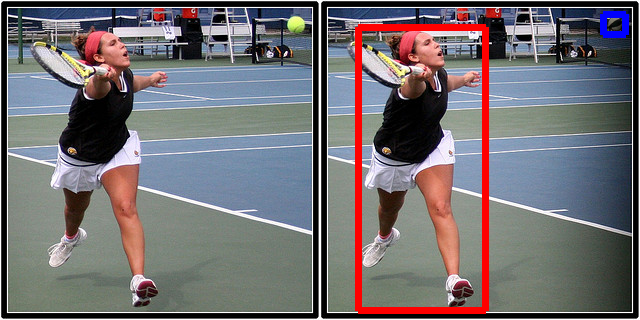}
\includegraphics[width=0.2\linewidth]{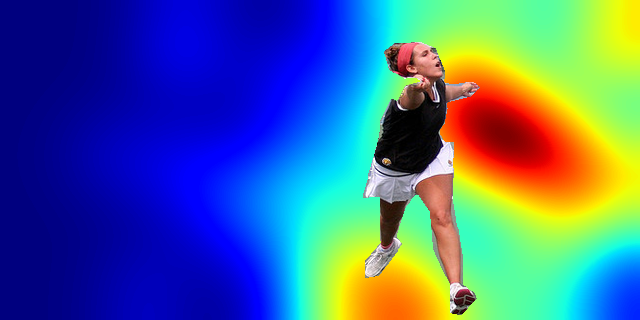}
\includegraphics[width=0.2\linewidth]{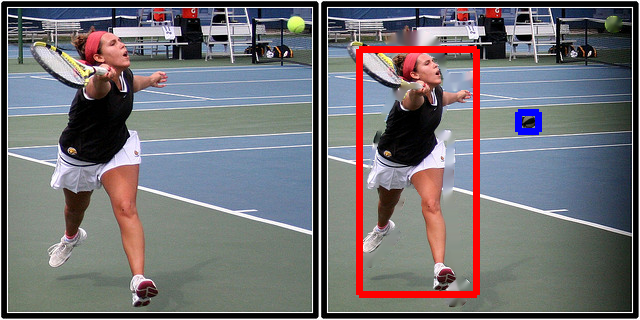}
}
\subfigure[ride the bicycle]{
\includegraphics[width=0.2\linewidth]{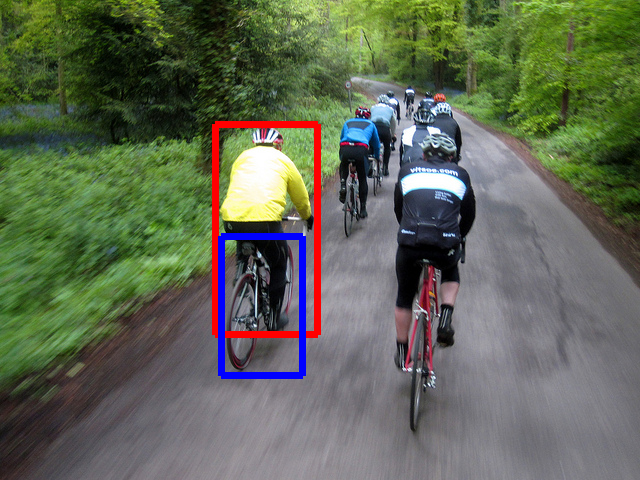}
\includegraphics[width=0.2\linewidth]{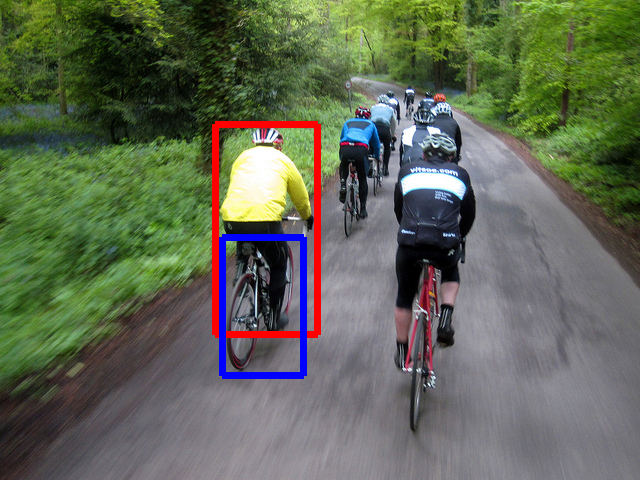}
\includegraphics[width=0.2\linewidth]{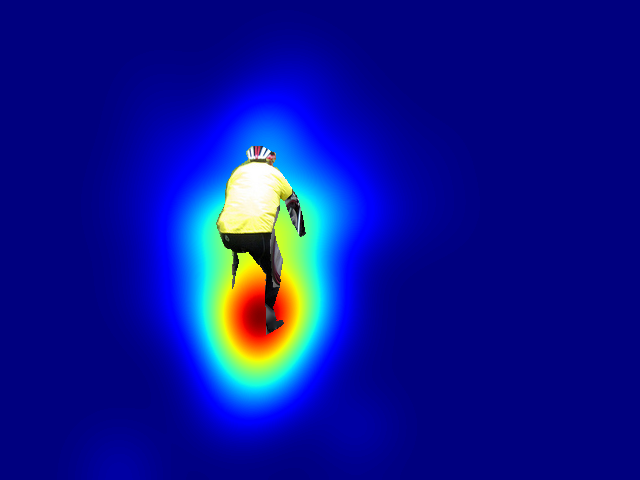}
\includegraphics[width=0.2\linewidth]{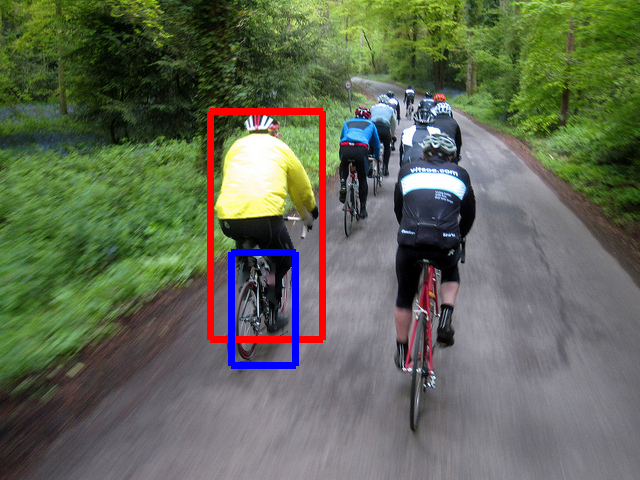}
}
\subfigure[carry the board]{
\includegraphics[width=0.2\linewidth]{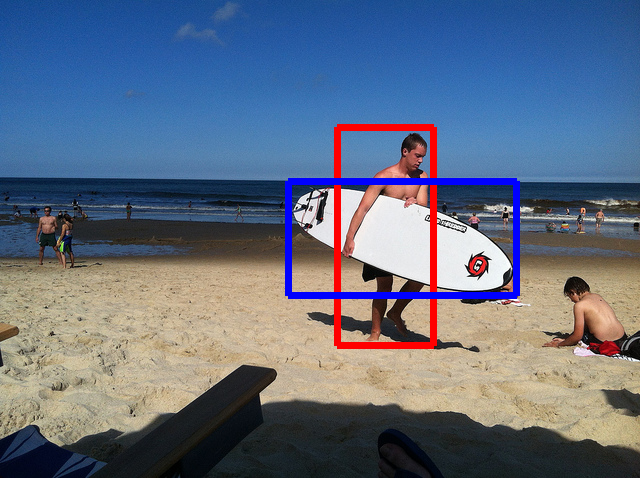}
\includegraphics[width=0.2\linewidth]{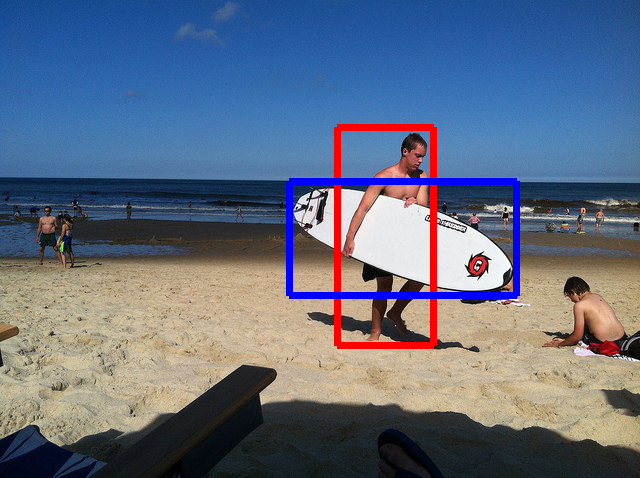}
\includegraphics[width=0.2\linewidth]{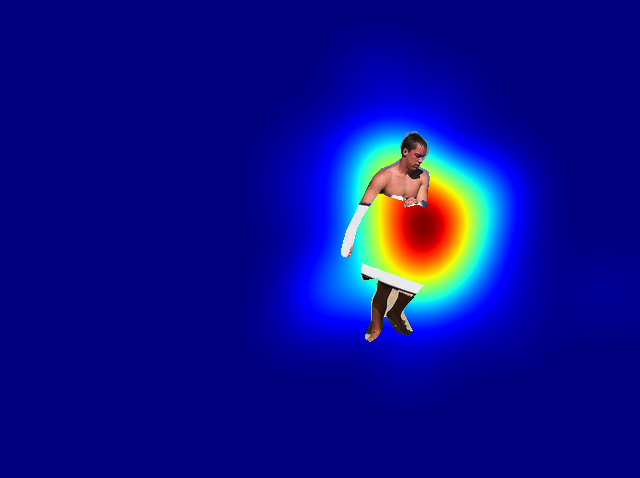}
\includegraphics[width=0.2\linewidth]{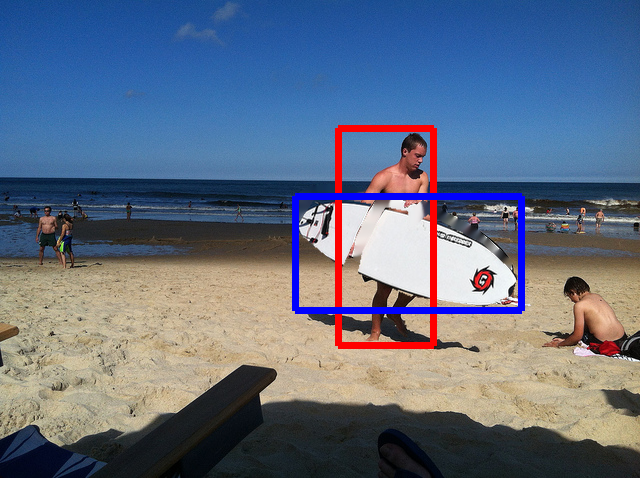}
}
\subfigure[stand on the board]{
\includegraphics[width=0.2\linewidth]{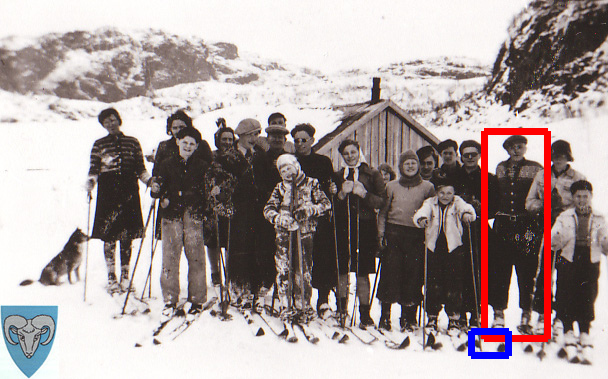}
\includegraphics[width=0.2\linewidth]{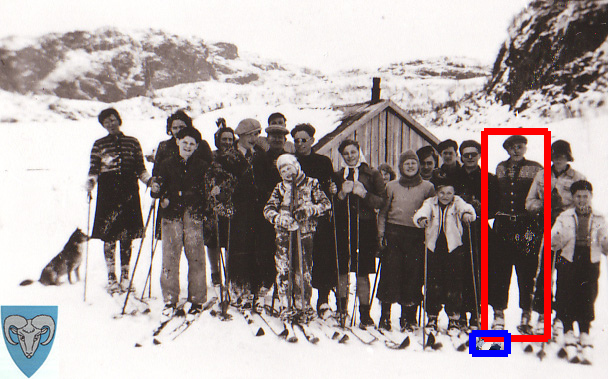}
\includegraphics[width=0.2\linewidth]{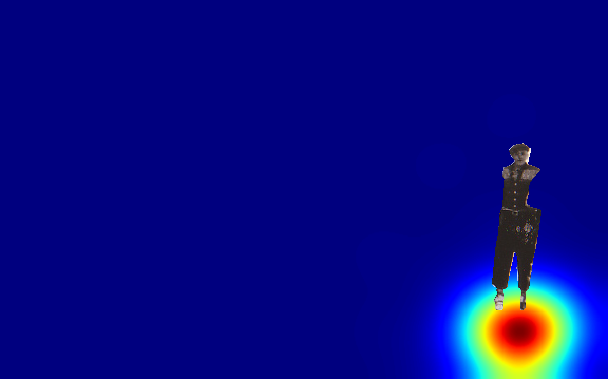}
\includegraphics[width=0.2\linewidth]{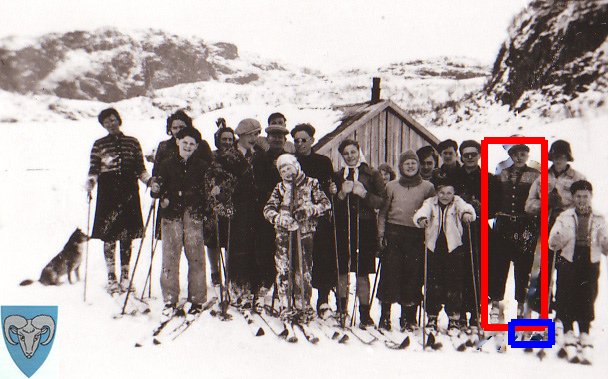}
}
\subfigure[hit/look the ball]{
\includegraphics[width=0.2\linewidth]{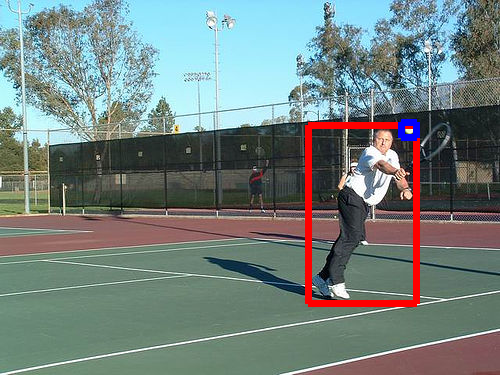}
\includegraphics[width=0.2\linewidth]{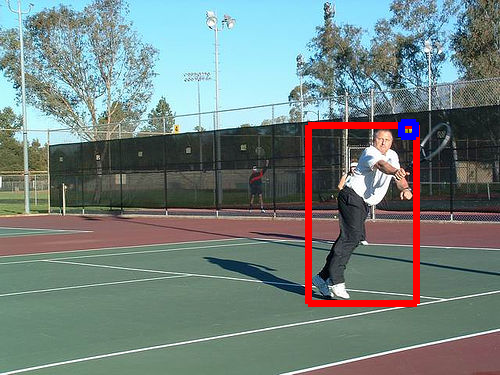}
\includegraphics[width=0.2\linewidth]{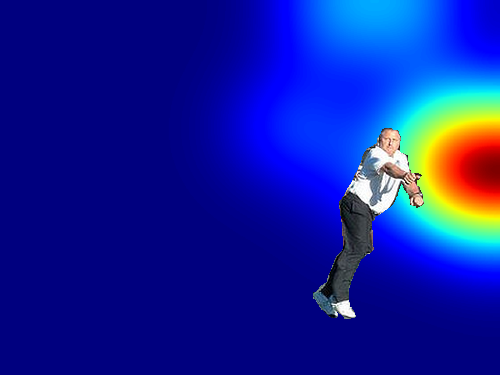}
\includegraphics[width=0.2\linewidth]{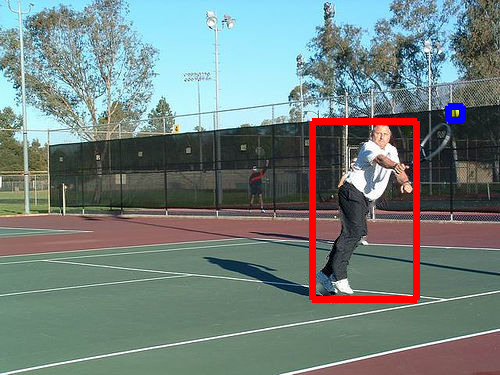}
}
\end{figure*}

\begin{figure*}
\centering
\subfigure[hold/look the phone]{
\includegraphics[width=0.2\linewidth]{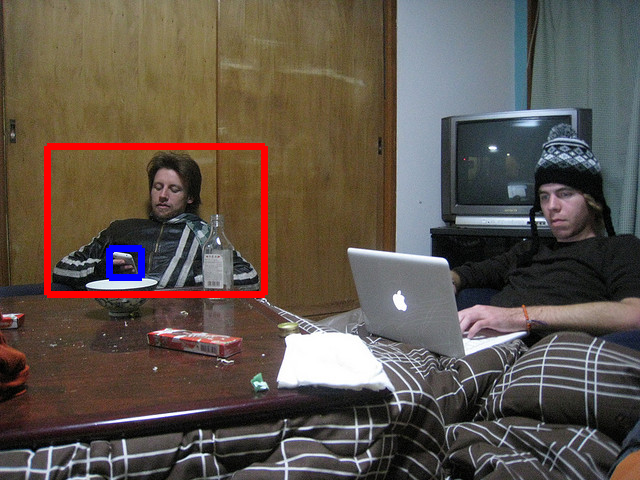}
\includegraphics[width=0.2\linewidth]{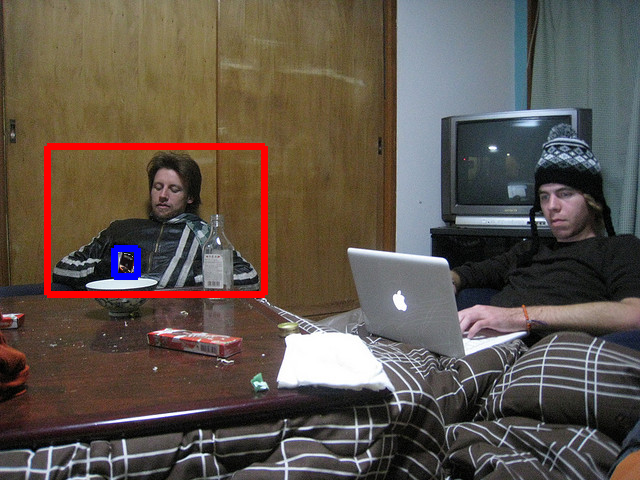}
\includegraphics[width=0.2\linewidth]{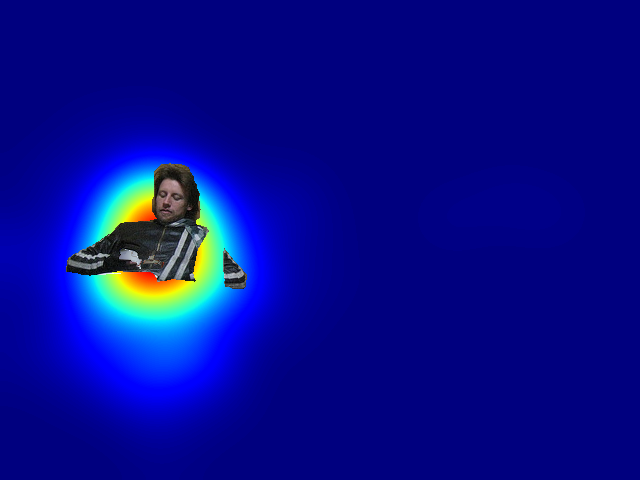}
\includegraphics[width=0.2\linewidth]{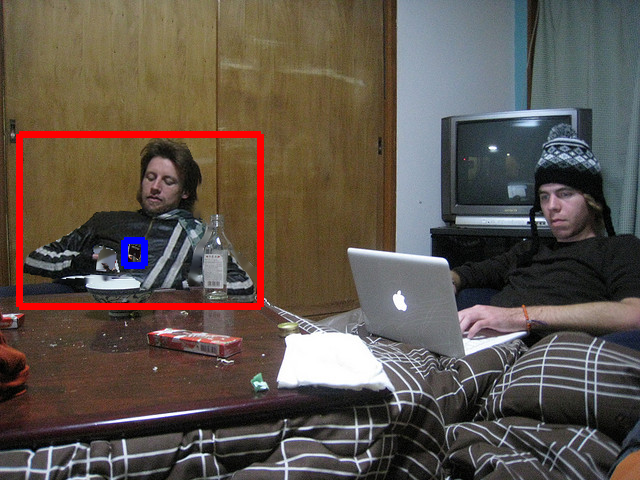}
}
\subfigure[snowboard with the board]{
\includegraphics[width=0.2\linewidth]{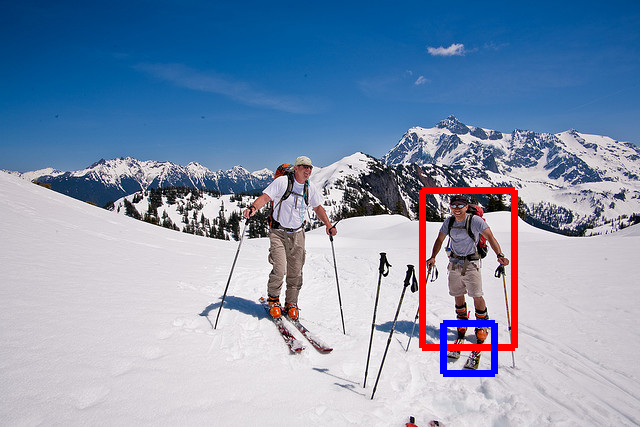}
\includegraphics[width=0.2\linewidth]{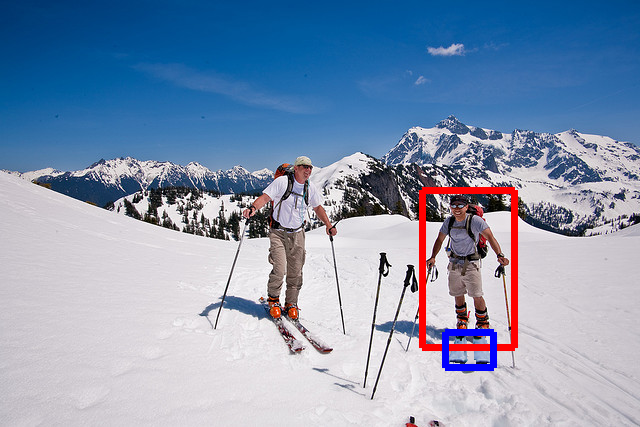}
\includegraphics[width=0.2\linewidth]{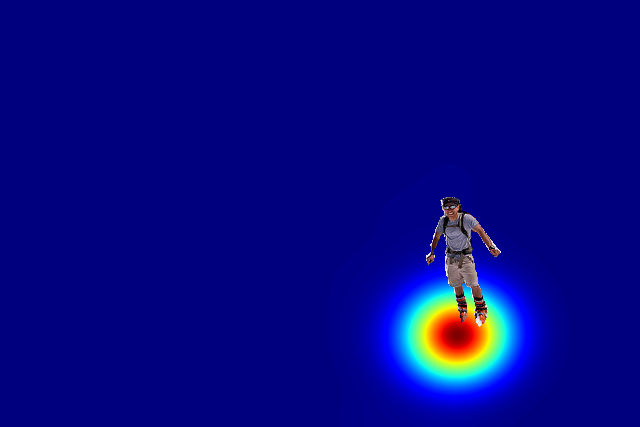}
\includegraphics[width=0.2\linewidth]{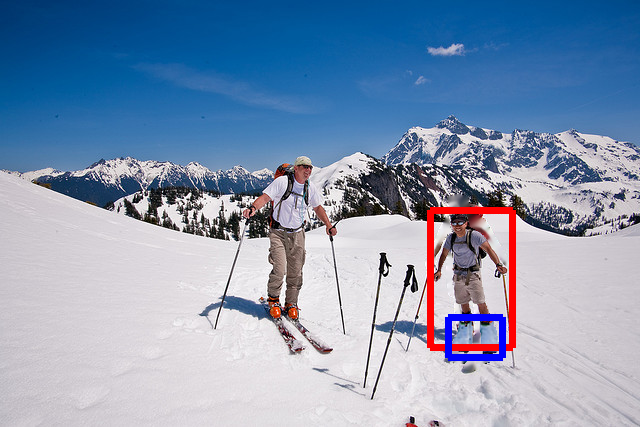}
}
\subfigure[look the kite]{
\includegraphics[width=0.2\linewidth]{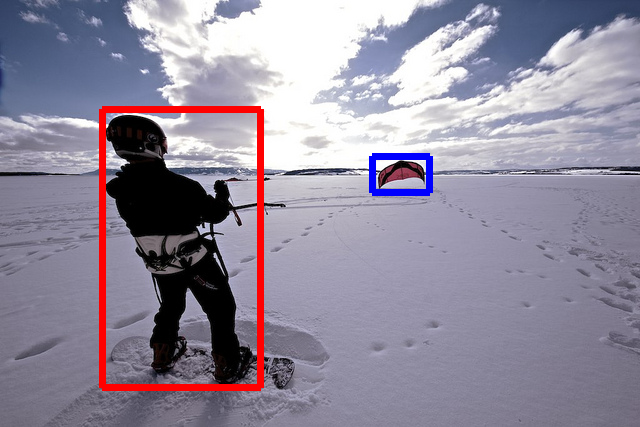}
\includegraphics[width=0.2\linewidth]{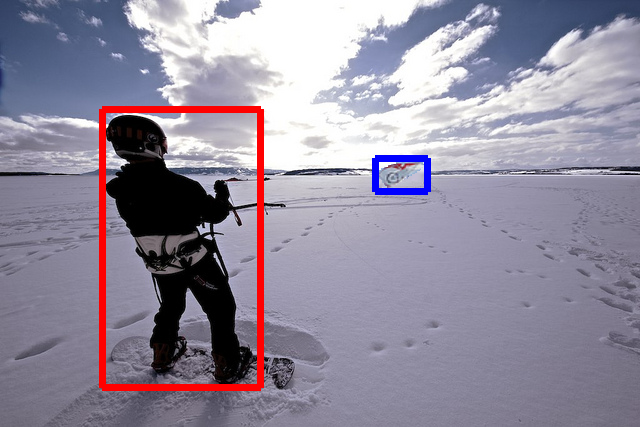}
\includegraphics[width=0.2\linewidth]{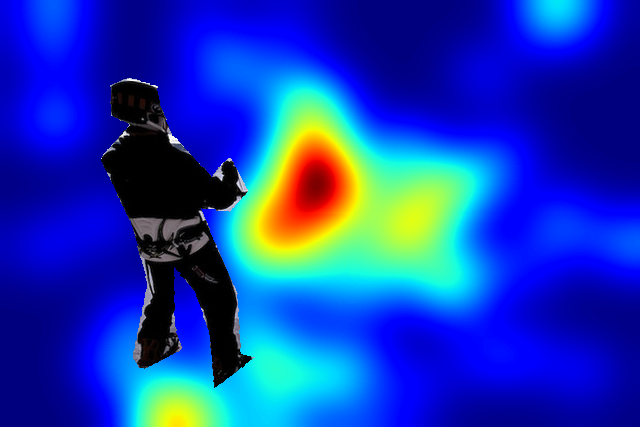}
\includegraphics[width=0.2\linewidth]{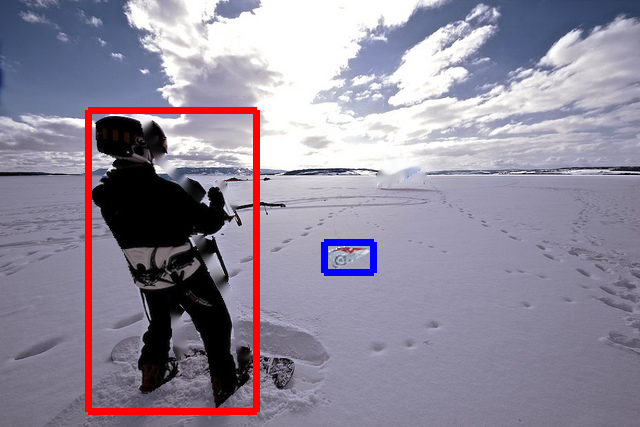}
}
\subfigure[ride the elephant]{
\includegraphics[width=0.2\linewidth]{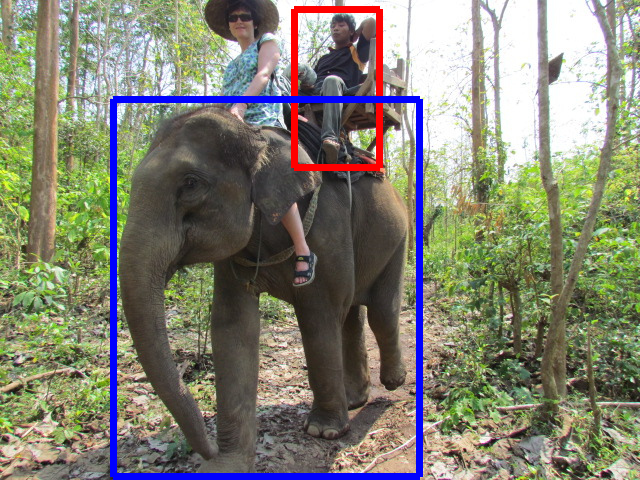}
\includegraphics[width=0.2\linewidth]{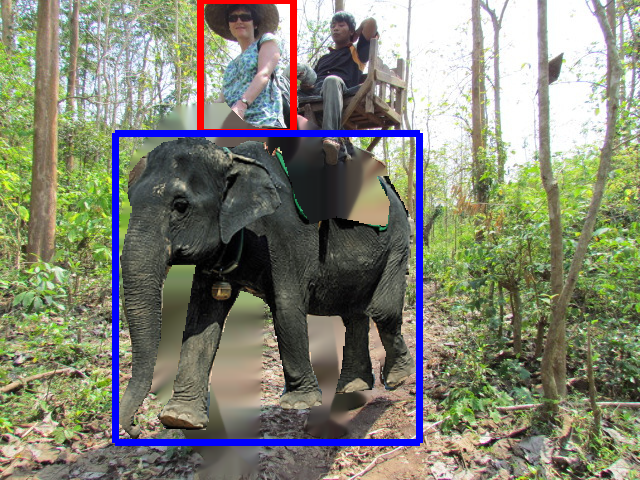}
\includegraphics[width=0.2\linewidth]{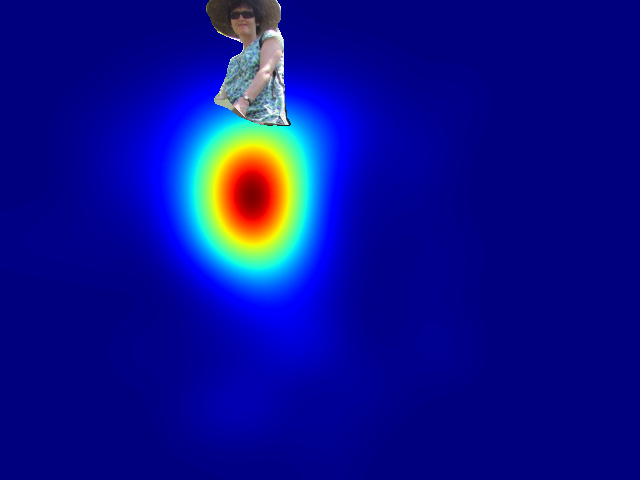}
\includegraphics[width=0.2\linewidth]{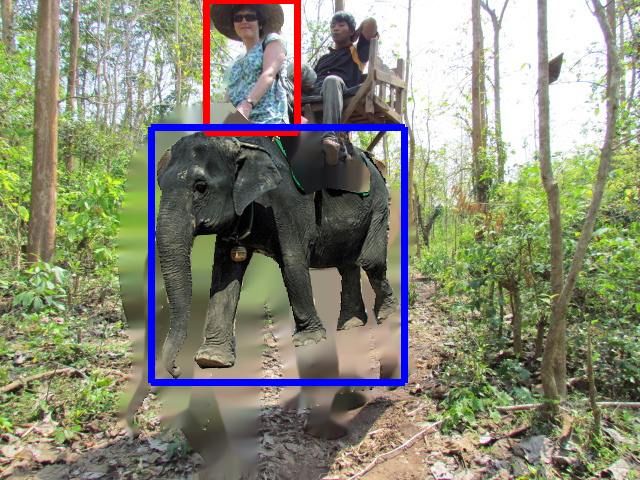}
}
\subfigure[look the frisbee]{
\includegraphics[width=0.2\linewidth]{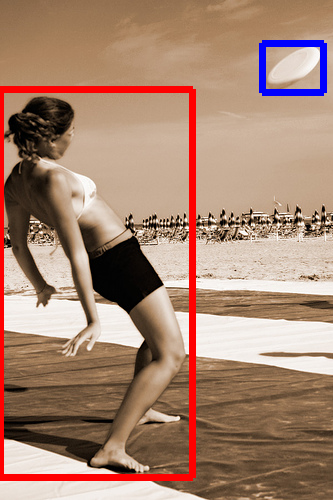}
\includegraphics[width=0.2\linewidth]{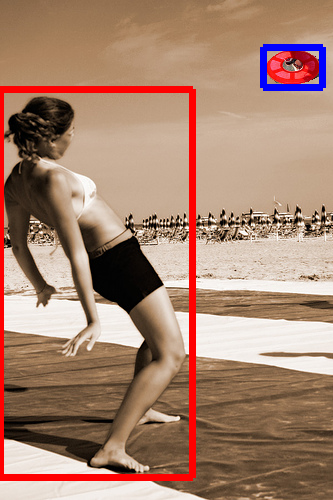}
\includegraphics[width=0.2\linewidth]{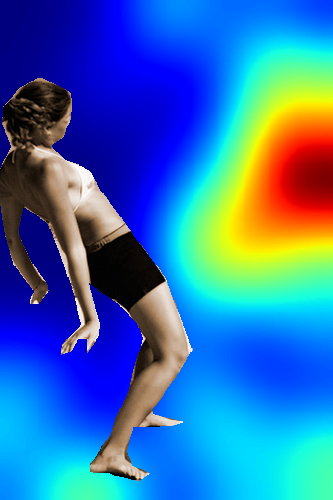}
\includegraphics[width=0.2\linewidth]{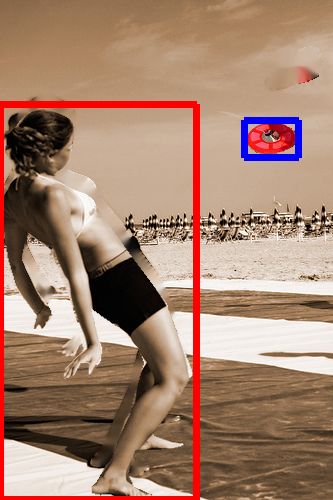}
}
\subfigure[skateboard with the board]{
\includegraphics[width=0.2\linewidth]{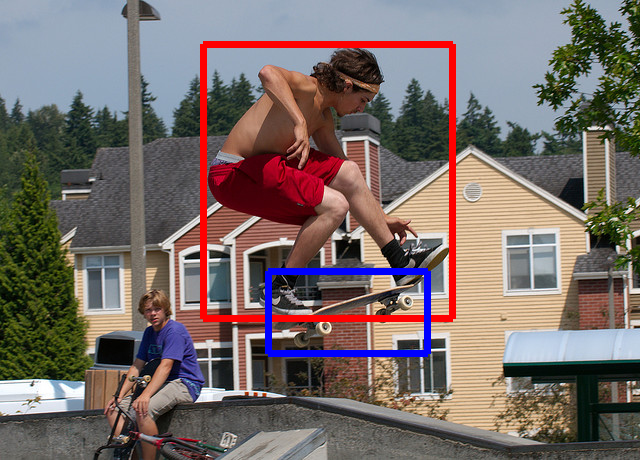}
\includegraphics[width=0.2\linewidth]{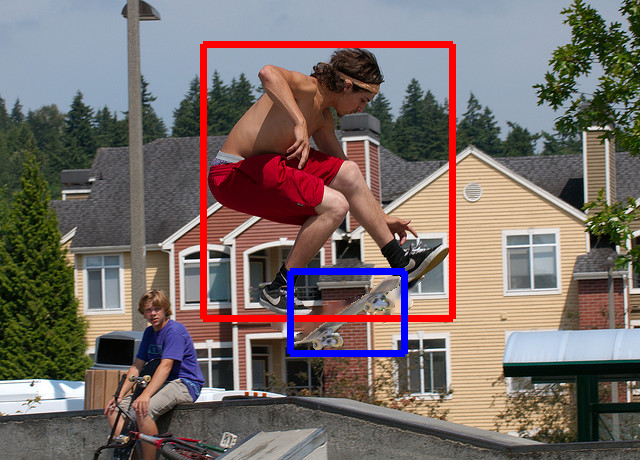}
\includegraphics[width=0.2\linewidth]{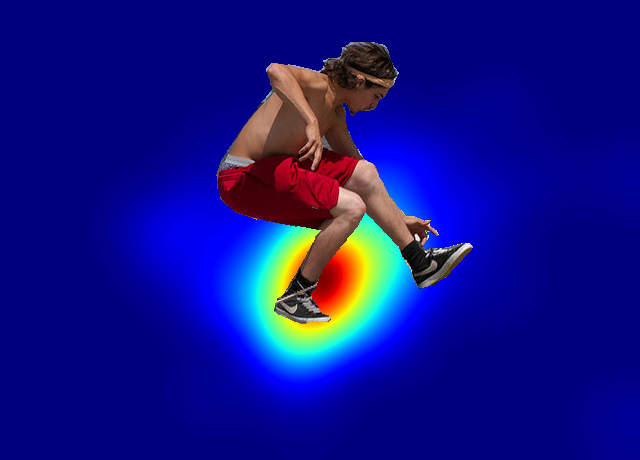}
\includegraphics[width=0.2\linewidth]{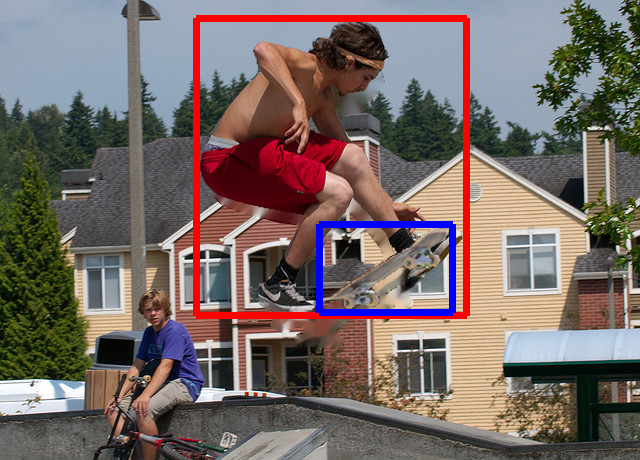}
}
\end{figure*}

\end{document}